\documentclass[Afour,sageh,times]{sagej}
\usepackage{moreverb,url}

\usepackage[colorlinks,bookmarksopen,bookmarksnumbered,citecolor=red,urlcolor=red]{hyperref}

\usepackage{multirow}
\usepackage{bm}
\usepackage{mathrsfs}
\usepackage{makecell}

\usepackage{soul}
\soulregister{\citep}7 
\soulregister{\ref}7 
\soulregister{\textcolor}7 

\usepackage{amsthm}
\newtheorem*{theorem}{Theorem}

\begin{document}

\runninghead{Liu et al}

\title{An eight-neuron network for quadruped locomotion with hip-knee joint control}

\author{Yide Liu\affilnum{1}, Xiyan Liu\affilnum{1}, Dongqi Wang, Wei Yang and Shaoxing Qu}

\affiliation{\affilnum{1}Yide Liu and Xiyan liu contributed equally to this work\\
All authors are with the State Key Laboratory of Fluid Power \& Mechatronic Systems, Key Laboratory of Soft Machines and Smart Devices of Zhejiang Provinces, Center for X-Mechanics, Department of Engineering Mechanics, Zhejiang University, Hangzhou 310027, China.}

\corrauth{Shaoxing Qu, Zhejiang University,
Hangzhou 310027,
China.}

\email{squ@zju.edu.cn}

\begin{abstract}
The gait generator, which is capable of producing rhythmic signals for coordinating multiple joints, is an essential component in the quadruped robot locomotion control framework.
The biological counterpart of the gait generator is the Central Pattern Generator (abbreviated as CPG), a small neural network consisting of interacting neurons.
Inspired by this architecture, researchers have designed artificial neural networks composed of simulated neurons or oscillator equations.
Despite the widespread application of these designed CPGs in various robot locomotion controls, some issues remain unaddressed, including:
(1) Simplistic network designs often overlook the symmetry between signal and network structure, resulting in fewer gait patterns than those found in nature.
(2) Due to minimal architectural consideration, quadruped control CPGs typically consist of only four neurons, which restricts the network's direct control to leg phases rather than joint coordination.
(3) Gait changes are achieved by varying the neuron couplings or the assignment between neurons and legs, rather than through external stimulation.
We apply symmetry theory to design an eight-neuron network, composed of Stein neuronal models, capable of achieving five gaits and coordinated control of the hip-knee joints.
We validate the signal stability of this network as a gait generator through numerical simulations, which reveal various results and patterns encountered during gait transitions using neuronal stimulation.
Based on these findings, we have developed several successful gait transition strategies through neuronal stimulations.
Using a commercial quadruped robot model, we demonstrate the usability and feasibility of this network by implementing motion control and gait transitions.
\end{abstract}


\maketitle

\section{Introduction}
Central pattern generator (CPG) is a small neural network composed of neurons with interactions~\citep{ijspeert_central_2008}, and it has been widely demonstrated to exist in the central nervous system of vertebrates ~\citep{grillner_central_1985,grillner_neurobiological_1985,grillner_vertebrate_1998} and the ganglia of invertebrates ~\citep{orlovsky_neuronal_1999}.
CPGs can generate primary signals for rhythmic behaviors such as locomotion and respiration without sensory feedback, while sensory feedback can be involved to shape the signals~\citep{yu_survey_2014}.

In robotics, CPGs are widely modeled and implemented in programs or hardware for controlling the locomotion and behavior of various types of robots, such as:
salamander robot for investigating the neural mechanisms behind salamander swimming and walking ~\citep{ijspeert_swimming_2007, ijspeert_amphibious_2020, thandiackal_emergence_2021},
integration with sensory feedback for bipedal  ~\citep{righetti_programmable_2006, vandernoot_bioinspired_2018}, quadrupedal ~\citep{righetti_pattern_2008, liu_cpginspired_2011}, and hexapod locomotion control ~\citep{manoonpong_modular_2007, manoonpong_sensordriven_2008, steingrube_selforganized_2010};
control of flapping-wing robot~\citep{chung_neurobiologically_2010, bayiz_experimental_2019} and fish robot~\citep{li_general_2015};
integration with reinforcement learning for controlling quadruped robots~\citep{shao_learning_2022, bellegarda_cpgrl_2022} and snake-like robots ~\citep{liu_reinforcement_2023},
and so on.

Compared with model-based control~\citep{dicarlo_dynamic_2018,kim_highly_2019} and learning-based control~\citep{tan_simtoreal_2018,lee_learning_2020}, the most attractive superiority of CPG is its simplicity and computational efficiency.
Previous works proved that the CPGs could be implemented on microcontroller~\citep{li_general_2015}, oscillating circuit on chips~\citep{still_neuromorphic_2006,saito_insecttype_2017,dutta_programmable_2019}, or even electronics-free pneumatic circuit~\citep{drotman_electronicsfree_2021}.
When serving as a gait generator in locomotion control, CPG possesses intrinsic stability and adaptability.
One only needs to compute a set of ordinary differential equations (ODEs) of the dynamical system to achieve coordinated rhythmic motion control of multiple joints of the robot.
These characteristics ensure the continuity of the locomotion and gait transition.

\begin{figure*}[ht]
  \centering
  \includegraphics[width = 16cm]{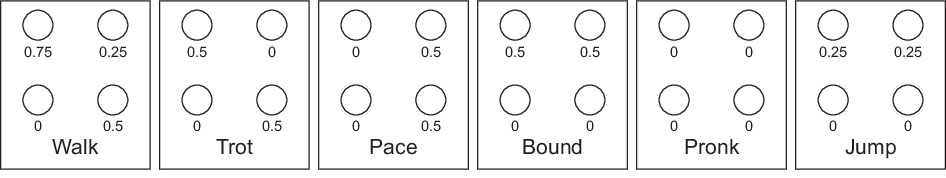}
  \caption{
  Phase relation for six types of quadruped locomotion gaits: walk, trot, pace, bound, pronk, and jump.
  In this work, the proposed eight-neuron network can achieve the above gaits except jump.
  }
  \label{fig:gaits_illustration}
\end{figure*}

While there are many projects involving quadruped robots using CPGs, the types of CPGs applicable to quadruped robots are limited.
Most gait generators for quadruped robots can be generalized as generating four primary signals through a four-neuron CPG, each corresponding to one of the four legs, to control the phase relationship among legs. As for hip-knee joint control, each primary signal undergoes a mapping function to generate signals that are independently assigned to knee and hip joints.
The type of CPG architecture based on a four-neuron network has many advantages, such as its simple structure, convenient tuning methods, ease of integration into a locomotion controller, and compatibility with combined approaches like sensor capability or learning-based control architecture.
However, it has several limitations as described below:

First, the number of gait rhythms that can be generated is limited.
It's observed in quadruped animals that there are many types of gaits. For instance, six primary gaits observed in animals~\citep{golubitsky_modular_1998,buono_models_2001} are walk, trot, pace, bound, pronk (or gallop), and jump (Fig.~\ref{fig:gaits_illustration}), while most of the four-neuron networks can achieve no more than three gait types.

Second, the number of joints/DoFs that can be controlled is limited.
The common design for quadrupeds is to have three joints per leg: hip abduction-adduction, hip flexion-extension, and knee flexion-extension.
However, in all four-neuron or even eight-neuron CPGs, only four signals can be utilized for phase relation control among four legs.
Realizing the phase relationship between the knee and hip joints often requires a mapping method to transfer the signal of a single neuron into two or more joint position signals.

For engineering applications, despite these limitations, a four-neuron network remains the most straightforward and effective architecture within the CPGs.
From a biological perspective, it is considered that control signals for multiple joints are likely produced by specific neurons rather than mapping functions ~\citep{golubitsky_modular_1998,righetti_control_2008}, necessitating more complex network architectures.
We believe that a CPG capable of generating various gaits and controlling multiple joints could offer new tools for gait generators in quadruped robots and stimulate research into new types of gaits and hip-knee coordinate control.
These visions motivate us to overcome the above limitations and design novel CPG network architectures.

A feasible routine to design CPG is to follow the symmetry principle. From a programming perspective, each neuron is a group of ODEs, which are also known as neuron models or oscillator models. The interrelations among neurons are named couplings, which can be represented by the coupling matrix. The symmetry of the CPG architecture is determined by the coupling matrix rather than the equation form of neurons. The whole CPG can be considered as a dynamic system, the rhythmic gaits correspond to the attractors of the system. The gait transition can be regarded as the system state switches from one attractor to the other.  The relationship between the symmetry of the network and the periodic solutions (gaits) has been revealed by the $H/K$ theorem, which will be further explained in the next section.
Thus, the solution for achieving gait variety is to design the symmetry variety of the network architecture. Moreover, the phase relationship between hip and knee joints should be maintained for all gaits, which means another type of intrinsic symmetry should be considered.

In this article, we designed an eight-neuron network for quadruped locomotion control with hip-knee joint control.
Here the hip joint refers to the hip flexion-extension DoF without abduction-adduction. The proposed eight-neuron network has three key features:
\begin{enumerate}
  \item {\it Gait diversity:}
  Due to its more complex network architecture, the eight-neuron network enables a wider range of gaits. In this study, we utilized the eight-neuron network to achieve five gaits: walk, trot, pace, bound, and pronk. The regulation of gaits can be achieved by simply modifying the control variables within the Mesencephalic locomotor region (MLR) module, its counterpart located in the midbrain of animals~\citep{cabelguen_bimodal_2003}.

  \item {\it Control more DoFs:}
  Most CPG with four or even eight neurons are designed to control the phase relations among legs, each motor's signal in a leg is obtained through post-processing by neurons.
  In engineering applications, this approach is more suitable for programmable foot trajectories.
  The eight-neuron network not only utilizes its first layer to accomplish similar tasks but also generates eight distinct control signals.
  Among these, every two neurons form a pair that can generate a group of signals directly assigned to both the knee and hip joints of a leg.
  This enables the achievement of phase relations among legs (i.e., gait) as well as phase relations between the hip and knee joints within each leg.
  These assignments are maintained during subsequent gait transitions.

  \item {\it Joint position continuity:}
  Since the neurons assigned to each joint remain unchanged, gait transition can be directly achieved by utilizing the MLR to regulate the eight-neuron network.
  We proposed four strategies by manipulating the variable in MLR to ensure the continuity and success of all twenty gait transitions.
\end{enumerate}

The architecture of the network has two four-neuron layers, one for hip joints and the other for knee joints.
Couplings of the network are designed based on the symmetry assumptions. The global symmetry refers to the phase relation among legs, the local symmetry refers to the phase relation between hip and knee joints in each leg.
Stein neuronal model~\citep{stein_improved_1974,stein_properties_1974} is modified as the ODEs of the neurons.
Control parameters for all gaits are verified by numerical simulation.
Gait transitions are investigated exhaustively. For certain gait transitions that were not encountered or difficult to ensure success in previous studies, we uncover the patterns between the outcomes of gait transitions and the transition time.
Based on these patterns, we propose several strategies to ensure successful transitions.
We further demonstrate the stability and simplicity of the proposed network as a gait generator for quadruped locomotion control through physical simulations on a commercial quadruped robot model.
The contributions of this work are:
\begin{enumerate}
  \item An eight-neuron network architecture for quadruped locomotion control.
  \item The realization of five gaits and transitions.
  \item The implementation of the network in the physical simulation.
\end{enumerate}

The organization of this paper is as follows.
In Section 2, we give an introduction to the symmetry theory of networks. We further propose the four-neuron network with $\mathbf{D}_4$ symmetry can generate five gaits.
In Section 3, we demonstrate the process of expanding the $\mathbf{D}_4$ four-neuron network into an eight-neuron network architecture. We also demonstrate the design of the ODEs of the neuron models.
Section 4 performs the numerical simulation results of the eight-neuron network, including the gaits and transitions. The patterns and strategies of gait transitions are demonstrated in detail.
Section 5 presents the physical simulation and the performance evaluation of the eight-neuron network by applying it to a commercial quadruped robot model.
In Section 6, we discuss the connections between the proposed network and previous researches, some assumptions made in the network design process, and potential improvements for further enhancing the proposed network.
Section 7 concludes this article.

\section{Symmetry of networks}
\subsection{Modeling CPG network}
From the perspective of biology, CPG is a small group of neurons that form a network. The connections among the neurons are called synapses. The behavior of the neuron can be modeled by the neuron models, which are generally a set of differential equations. The connections among the neurons can be modeled by the coupling effects, which can be further described by the coupling matrix. Modeling a CPG network can be separated into two steps:
\begin{enumerate}
  \item Design the network architecture.
  \item Design the neuron model.
\end{enumerate}

The architecture of the network can be presented as a graph with nodes and edges. The nodes represent the state variables and the edges represent the interactions among these variables.
CPGs are dynamic networks, the nodes represent the neurons and the edges represent the coupling effect of the neurons.
Here we give a brief example. As in ~\cite{collins_hardwired_1994}, the CPGs adopted the Hopf model, Van der Pol model, and Stein neuronal model, all these networks have the same architecture as shown in Fig.~\ref{fig:4neuron}(b).
The coupling with arrows forms a unidirectional ring.
An interesting phenomenon is that, despite the neurons of the CPGs being calculated by a variety of models, all of these CPGs are capable of generating walk, trot, and bound gaits.
Additionally, plenty of robotic research proved that the neurons of the CPGs are selected by a variety of forms of ODEs. Some of them are derived from the neurons' behavior, and some of them are simply oscillator equations. These CPGs can perform similar gait types as long as they share the same network architecture.

This can be explained as the symmetries of the network leading to the synchrony and phase relation of the neurons. The pattern of the network is determined by the architecture (or the symmetry) of the network, rather than ODEs. Thus, to design the CPG to achieve the desired gaits, the first step is to design the architecture of the network which determines the symmetries.

\subsection{Symmetries and $H/K$ theorem}
\label{section-2-B}
The symmetry in CPG refers to the transformation that preserves the states of the system. In CPGs, two terms of symmetries are considered: the symmetry of the network and the symmetry of the gait rhythm.
The symmetries of the network can be defined as a group of permutations that preserve network architecture.
The gait rhythm corresponds to the periodic states of each joint in each leg (such as the angles of the joint).
A brief introduction of the involved symmetry is given in Appendix B.

The $H/K$ theorem~\citep{buono_models_2001,golubitsky_nonlinear_2006} is introduced here to demonstrate the relationship between the gait rhythms and the symmetry of the network.
The theorem is stated as:
\begin{theorem}
  {($H/K$ Theorem)}
Let $\Gamma$ be the symmetry group of a coupled neuron network in which all neurons are coupled and the internal dynamics of each neuron is at least two-dimensional. Let $K\subset H \subset \Gamma$ be a pair of subgroups. Then there exist periodic solutions to some coupled neuron systems with spatiotemporal symmetries H and
spatial symmetries $K$ if and only if $H/K$ is cyclic and $K$ is an isotropy subgroup. Moreover, the system can be chosen so that the periodic solution is asymptotically stable.
\end{theorem}

\begin{table*}[t]
\small\sf\centering
\caption{Spatiotemporal group $H$, Subgroup $K$ and quotient subgroup $H/K$ in $\mathbf{D}_4$ for five gaits.}
\label{T:subgroup}
    \begin{tabular}{ccccccc}
        \toprule
        Gait type& $H$ & $K$ & $H/K$ & $x_2(t)$ & $x_3(t)$ & $x_4(t)$\\
        \midrule
       Walk & $\mathbf{Z}_4(\omega)$ & $\mathbf{1}$ & $\mathbf{Z}_4(\omega)$ & $x_1(t+\frac{1}{2})$ & $x_1(t+\frac{1}{4})$ & $x_1(t+\frac{3}{4})$\\

       \cmidrule(lr){1-7}

       Trot& $\mathbf{D}_2(k,\omega^2)$ & $\mathbf{Z}_2(k)$ & $\mathbf{Z}_2(\omega^2)$ & $x_1(t+\frac{1}{2})$ & $x_1(t)$ & $x_1(t+\frac{1}{2})$\\

       \cmidrule(lr){1-7}

       Pace& $\mathbf{D}_2(k,\omega^2)$ & $\mathbf{Z}_2(k\omega^2)$ & $\mathbf{Z}_2(k)$ & $x_1(t+\frac{1}{2})$ & $x_1(t+\frac{1}{2})$ & $x_1(t)$\\

       \cmidrule(lr){1-7}

       Bound& $\mathbf{D}_2(k,\omega^2)$ & $\mathbf{Z}_2(\omega^2)$ & $\mathbf{Z}_2(k)$ & $x_1(t)$ & $x_1(t+\frac{1}{2})$ & $x_1(t+\frac{1}{2})$\\

       \cmidrule(lr){1-7}

       Pronk& $\mathbf{D}_4(k,\omega)$ & $\mathbf{D}_4(k,\omega)$ & $\mathbf{1}$ & $x_1(t)$ & $x_1(t)$ & $x_1(t)$\\

       \bottomrule
    \end{tabular}

\end{table*}

\begin{figure}[t]
  \centering
  \includegraphics[width = 6cm]{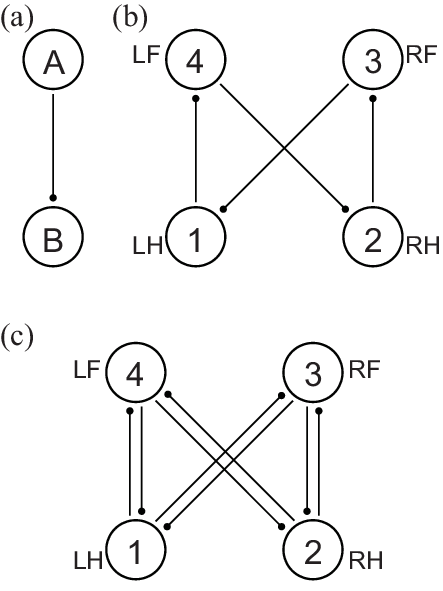}
  \caption{
  (a) The arrow represents the coupling effect from neuron A to neuron B.
  (b) The four-neuron network with one way-coupling. The graph-theoretic automorphism group is $\mathbf{Z}_4$.
  (c) The four-neuron network with two way-coupling. The graph-theoretic automorphism group is $\mathbf{D}_4$.
  Neurons 1, 2, 3, and 4 are assigned to control the left hind (LH), right hind (RH), right front (RF), and left front (LF) limbs of the quadruped, respectively.
  }
  \label{fig:4neuron}
\end{figure}

Here, we make some brief notes for this theorem.
\begin{itemize}
  \item CPG is considered as a network of neurons with couplings, which can generate gait rhythms.
  \item The symmetry of the CPG architecture is noted as group $\Gamma$.
  \item The gait rhythms are time domain signals with two types of symmetry, the spatial and the temporal symmetries. The spatiotemporal symmetry of a specific gait is noted as group $H$.
  \item Proving that a CPG architecture can generate a type of gait involves three steps:
  (1) Derive the symmetry group of the network $\Gamma$.
  (2) Derive the spatiotemporal symmetry group $H$ of the gait, which should be a subgroup in $\Gamma$.
  (3) Derive the spatial symmetry group $K$ of the gait, which should be a subgroup in $H$. $K$ is an isotropy group, which corresponds to the spatial symmetry of the gait.
  (4) Calculate the quotient group $H/K$. $H/K$ should be a cyclic group. $H/K$ corresponds to the temporal symmetry of the gait.
\end{itemize}

For example, in classical CPG architecture $\mathbf{Z}_4$ (Fig.~\ref{fig:4neuron}(b)),
neurons 1, 2, 3, and 4 are assigned to control the left hind (LH), right hind (RH), right front (RF), and left front (LF) limbs of the quadruped, respectively.
The group generator is $\omega=(1324)$, there are four elements:
\begin{equation}\label{eq:0}
  \mathbf{Z}_4=\{e,\omega,\omega^2,\omega^3\}.
\end{equation}
For walk, trot, and bound gait in ~\cite{collins_hardwired_1994}, the $H$ and $K$ combinations are derived as:
\begin{itemize}
  \item \textbf{Walk}:
  The $H$, $K$, and $H/K$ are $\mathbf{Z}_4(\omega)$, $e$, and $\mathbf{Z}_4(\omega(\frac{1}{4}))$, respectively. The $\frac{1}{4}$ refers to the phase shift generated by the $H/K$ cyclic group.

  \item \textbf{Trot}:
  The $H$, $K$, and $H/K$ are $\mathbf{Z}_2(\omega^2)$, $e$, and $\mathbf{Z}_2(\omega^2(\frac{1}{2}))$.

  \item \textbf{Bound}:
  The $H$, $K$, and $H/K$ are  $\mathbf{Z}_4(\omega)$, $\mathbf{Z}_2(\omega^2)$, and $\mathbf{Z}_2(\omega(\frac{1}{2}))$.
\end{itemize}

\subsection{Four-neuron network with $\mathbf{D}_4$ symmetry}
We aim to build a CPG for quadruped locomotion with multiple gaits: walk, trot, pace, bound, and pronk. Previous work~~\citep {golubitsky_nonlinear_2006} proved that in a network with $\mathbf{Z}_4$ symmetry, the symmetry group of trot and pace were always conjugate.
This conclusion implies that the trot and pace gaits are unable to coexist in a single $\mathbf{Z}_4$ network, unless the coupling is modified~\citep{song_gaits_2023}.
Thus, we consider the symmetry of the network as {$\mathbf{D}_4$}, which has more elements than {$\mathbf{Z}_4$}. The corresponding network architecture is a four-neuron network with two-way coupling (Fig.~\ref{fig:4neuron}(c)).
The generator of the $\mathbf{D}_4$ group is $\omega = (1324)$ and $k=(13)(24)$. The elements of the $\mathbf{D}_4$ can be expressed as:

\begin{equation}\label{eq:1}
  \mathbf{D}_4=\{e,\omega,\omega^2,\omega^3,k,k\omega,k\omega^2,k\omega^3\}.
\end{equation}

According to the $H/K$ theorem, we need to find the appropriate cyclic quotient subgroup and isotropy subgroup of the $\mathbf{D}_4$  that conforms to the gaits we need.
For five desired gaits, the phase relation is illustrated in Fig.~\ref{fig:gaits_illustration}, the derivation process is demonstrated below:

\begin{itemize}
  \item \textbf{Walk}:
  The cyclic subgroup $H/K$ can be derived from the phase relation among the legs, that is $\mathbf{Z}_4(\omega(\frac{1}{4}))$.

  \item \textbf{Trot}:
  The spatial symmetry of the trot gait is diagonal equivalence, which implies the isotropic subgroup $K$ of the trot gaits is $\mathbf{Z}_2(k)$.
  The phase relation among the leg is $\frac{1}{2}$, thus the $H/K$ is calculated as $\mathbf{Z}_2(\omega^2(\frac{1}{2}))$.

  \item \textbf{Pace}:
  The spatial symmetry maintains the equivalence between neurons 1,4 and neurons 2,3.
  The isotropic subgroup of the pace is selected $K=\mathbf{Z}_2(k\omega^2)$.
  The phase relation among the diagonal legs is  $\frac{1}{2}$, which corresponds to $H/K=\mathbf{Z}_2(k(\frac{1}{2}))$.

  \item \textbf{Bound}:
  The spatial symmetry maintains the equivalence between neurons 1,2 and neurons 3,4.
  The isotropic subgroup of the bound is selected as $K=\mathbf{Z}_2(\omega^2)$.
  The phase relation among the diagonal legs is  $\frac{1}{2}$,  which corresponds to $H/K=\mathbf{Z}_2(k(\frac{1}{2}))$.

  \item \textbf{Pronk}:
  All neurons are in phase, and the isotropic subgroup of the spatial symmetry is $K=\mathbf{D}_4(k,\omega)$.
  $H/K=e$.
\end{itemize}

The subgroup combinations of the above gaits are listed in Table \ref{T:subgroup}. The symbol $x_i$ represents the states of the neuron $i$, and the phase relations of other neurons compared with neuron 1 in different gaits are also listed. It is proved that the four-neuron network with symmetry $\mathbf{D}_4$  meets the requirements of the spatiotemporal symmetries of all desired gaits. The symmetry of our network is set as $\mathbf{D}_4$ for the following design process.

\section{Eight-neuron network and Stein neuronal model}
Based on the foundation that the four-neuron $\mathbf{D}_4$ network with two-way coupling can generate five types of gaits, this section further introduces how to expand the network architecture from four-neurons to eight-neurons.

Most quadruped CPGs employ four-neuron networks, and the signals generated by neurons are assigned to the hip joints correspondingly. The signals of knee joints are generated by mapping functions based on the designed trajectories.
The four-neuron network has its advantages, the simple architecture and computational efficacy allow it to be implemented into the locomotion controller easily.

Designing a network architecture more complex than the four-neuron architecture does not compromise these advantages. A network with additional neurons can have more symmetries, which may generate more gait types. Moreover, achieving hip-knee coordination through the inherent characteristics of the network rather than manually designed trajectories can offer more insights into gait transition mechanisms and performance, benefiting both robotics and biology.

In this work, we aim to build a CPG to achieve hip-knee control of the quadrupeds. The neurons of the network can generate signals for both the hip and knee joints of the robot. Each joint has an assigned neuron.
Based on the gait existence proof of the four-neuron $\mathbf{D}_4$ network, we can expand it to eight neurons.
The two challenges that exist here are:
\begin{enumerate}
  \item Adding neurons while maintaining the $\mathbf{D}_4$ symmetry.
  \item Designing symmetry to achieve the phase locking between the hip and knee neurons.
\end{enumerate}

Thus far, we conclude that designing an eight-neuron network for hip-knee coordinated control requires consideration of two types of symmetry. One corresponds to the gait rhythm (among four legs), while the other corresponds to the phase relationship of the hip-knee joints. In subsequent sections, the former is referred to as global symmetry, while the latter is termed local symmetry.

\begin{figure}[t]
	\begin{minipage}{1\linewidth}
  \centering
  \includegraphics[width=8cm]{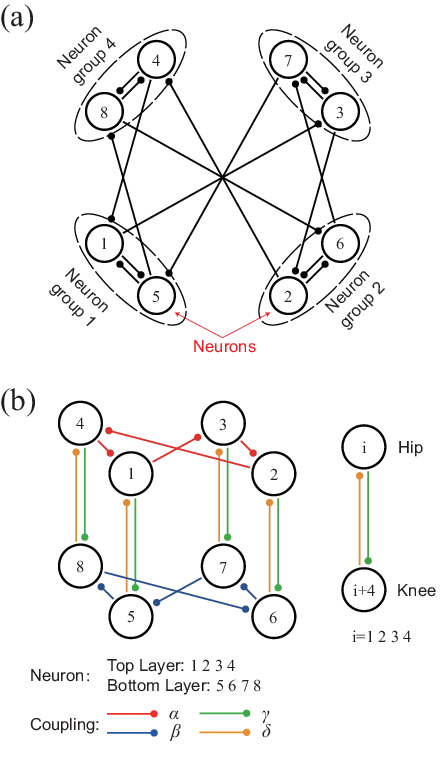}
  \caption{
  (a) Consider each neuron in $\mathbf{D}_4$ four-neuron network as a small group. Each group is divided into two neurons with two-way couplings. The network is thus expanded into eight neurons. Both global and local symmetry are maintained.
  (b) The cube architecture of the eight-neuron network. The neurons in the top and bottom layers correspond to the hip and knee joints of a leg. The network has four types of couplings: $\alpha$, $\beta$, $\gamma$ and $\delta$.
  }
  \label{fig:8neuron}
	\end{minipage}

\begin{minipage}{0.48\textwidth}
\small\sf\centering
\vspace{0.5 cm}
\captionof{table}{Coupling matrix $\lambda_{ij}$ of the eight-neuron network.}
\label{T:coupling matrix}
    \begin{tabular}{c|cccc|cccc}
    Neuron& 1 & 2 & 3 & 4 & 5 & 6 & 7 & 8\\
    \cmidrule{1-9}
      1 & $0$ & $0$ & $1$ & $0$ & $1$ & $0$ & $0$ & $0$ \\
      2 & $0$ & $0$ & $0$ & $1$ & $0$ & $1$ & $0$ & $0$ \\
      3 & $0$ & $1$ & $0$ & $0$ & $0$ & $0$ & $1$ & $0$ \\
      4 & $1$ & $0$ & $0$ & $0$ & $0$ & $0$ & $0$ & $1$ \\
      \cmidrule{1-9}
      5 & $1$ & $0$ & $0$ & $0$ & $0$ & $0$ & $0$ & $1$ \\
      6 & $0$ & $1$ & $0$ & $0$ & $0$ & $0$ & $1$ & $0$ \\
      7 & $0$ & $0$ & $1$ & $0$ & $1$ & $0$ & $0$ & $0$ \\
      8 & $0$ & $0$ & $0$ & $1$ & $0$ & $1$ & $0$ & $0$
    \end{tabular}

\end{minipage}
\end{figure}

\subsection{Expand four-neuron network to eight-neuron network}
There are various approaches to expand a four-neuron network to the one composed of eight neurons. One straightforward method is adding four neurons and building couplings. However, this approach could potentially impact the global symmetry of the original network, the local symmetry is also difficult to design this way.

In this paper, we increase the number of neurons in the network by splitting one neuron in the $\mathbf{D}_4$ network into two neurons (Fig.~\ref{fig:8neuron}(a)).
We take the foundation of the four-neuron $\mathbf{D}_4$ network and split one existing neuron into a sub-group consisting of two neurons. Within this group, the two neurons are connected bidirectionally, establishing local symmetry $\mathbf{Z}_2$, while preserving the original global symmetry. In an expanded eight-neuron network, the generator corresponding to local symmetry is $\lambda=(15)(26)(37)(48)$, resulting in the group generator for the eight-neuron network being:

\begin{equation}
      \begin{aligned}\label{eq:generator difference}
        \omega & =(1324)(5768) \\
        \kappa & =\lambda \cdot (13)(24)(57)(68)=(17)(35)(28)(46)
\end{aligned}
\end{equation}

The expanded eight-neuron network architecture can be further represented by a 3-D cube, which is shown in Fig.~\ref{fig:8neuron}(b). The original neuron assigned to a leg in $\mathbf{D}_4$ four-neuron network is divided into two neurons with two-way couplings. The upper is for the hip joint and the bottom is for the knee joint.
The neuron pairs 1 and 5 are assigned to control the hip and knee of the left hind leg of the quadruped, and the neuron pairs 2,6 for the right hind, 3,7 for the right front, and 4,8 for the left front, respectively.

The two-way couplings in $\mathbf{D}_4$ four-neuron network are also separated into two layers, the top layer and the bottom layer both contain a one-way coupling ring but in different directions. New couplings are added between the top and bottom neurons to maintain the equivalence of the hip and knee neurons. The local symmetry formed by the couplings between neuron pairs can be derived from adding a self-coupling for each neuron in the $\mathbf{D}_4$ four-neuron network~\citep{golubitsky_patterns_2005} .

Considering the couplings between the top and bottom layer, there are four types of couplings in Fig.~\ref{fig:8neuron}(b):
\begin{itemize}
  \item $\alpha$: the coupling in the top layer
  \item $\beta$: the coupling in the bottom layer
  \item $\gamma$: the coupling from top to the bottom
  \item $\delta$: the coupling from the bottom to the top
\end{itemize}

To maintain the $\mathbf{D}_4$ global symmetry, couplings $\alpha$ and $\beta$ should be equivalent, leading to the equivalence of the neurons in the same layer. Moreover, the neuron pair for a leg is also equivalent, leading to the equivalence of the couplings $\gamma$ and $\delta$.
The local symmetry is formed by the $\gamma$ and $\delta$. So far, we still consider they are equal, further modification will be discussed in subsequent sections.

\subsection{Stein neuronal model and modifications}
In CPG design, the models of the neuron can be classified as human-designed oscillators and biological neuron models.
The advantage of the oscillators is their relatively simple form, and the limit cycles are adjustable.
The biological neuron models are typically proposed by neurobiologists.
The equations of these models are relatively complex and can simulate certain behaviors of neurons.
The parameters in the models often have clear biological and physical meanings.

In this work, we employ the Stein model, a biological neuron model first introduced in \cite{stein_improved_1974, stein_properties_1974}.
We chose this model because it effectively describes the pulse and step responses of neurons, and the parameters in the model have clear physical interpretations.
The equations of this model can be represented as:
\begin{equation}\label{eq:stein model}
    \begin{aligned}
    \dot{x_i} &=a[-x_i+\frac{1}{1+exp(-f_{ci}-by_i+bz_i)}] \\
    \dot{y_i} &=x_i-py_i\\
    \dot{z_i} &=x_i-qz_i\\
    \end{aligned}
\end{equation}
where $x_i$ is the membrane potential of the $i$th neuron, $a$ is a rate constant affecting the frequency of the neuron, in the eight-neuron network with two layers, there are parameters $a^h$ and $a^k$, correspondingly. The superscripts $h$ and $k$ refer to hip and knee joints. $f_{ci}$ is the driving signal for this neuron. $b$ is the self-adaption constant which determines the extent of the adaptation. $p$ and $q$ are the rate constants of the transitions referring to the sodium ionic accumulation~\citep{hodgkin_potassium_1960}. In this work, the constants are set as $b=-2000, p=10, q=30$.

In the eight-neuron network, the coupling effects from the other neurons are included in the term of the driving signal $f_{ci}$.
The $f_{ci}$ has three components: control signals for gait selection and transition from the MLR, coupling effects from the neurons in the same layer, and coupling effects from the neurons in the other layer.
In this work, $f_{ci}$ are expressed as:

\begin{equation}\label{eq:driving signals}
    \begin{aligned}
    f_{ci}^h = & f^h[1+k_1^h\sin(k_2^ht)+\alpha \sum_{j=1}^{4}\lambda_{ji}x_j+\delta \sum_{j=5}^{8}\lambda_{ji}x_j]\\
    &\text{For top layer:} i=1,2,3,4\\
    f_{ci}^k = & f^k[1+k_1^k\sin(k_2^kt)+\beta \sum_{j=5}^{8}\lambda_{ji}x_j+\gamma \sum_{j=1}^{4}\lambda_{ji}x_j]\\
    &\text{For bottom layer:} i=5,6,7,8\\
    \end{aligned}
\end{equation}
where $f(f^h, f^k)$ is an amplitude parameter, $k_1(k_1^h, k_1^k)$ and $k_2(k_2^h, k_2^k)$ determine the amplitude and frequency of the driving signal from the MLR. $\alpha \sum_{j=1}^{4}\lambda_{ji}x_j$ and $\beta \sum_{j=5}^{8}\lambda_{ji}x_j$ refer to the coupling effects from the neurons in the same layer,  $\delta \sum_{j=5}^{8}\lambda_{ji}x_j$ and $\gamma \sum_{j=1}^{4}\lambda_{ji}x_j$ refer to the coupling effects from the neurons in the other layer. Parameters $a, f, k_1$ and $k_2$ control the gaits of the network.

\subsection{Design couplings and model parameters}
\label{sec:Design couplings}
The coupling parameters are set based on the network architecture and the neuron model. The coupling matrix $\lambda_{ij}$ can be derived from the network architecture (Fig.~\ref{fig:8neuron}(b)) and is listed in Table~\ref{T:coupling matrix}.

Following the global symmetry, the coupling parameters $\alpha$ and $\beta$ are both set as -0.15, representing inhibitory couplings among neurons in the same layer.
As for local symmetry between neuron pairs, if $\gamma=\delta$, it will lead to a $\mathbf{Z}_2$ symmetry of the local neuron pair network.
Even though a pair of hip and knee neurons are not strictly equivalent since they lie in different layers with opposite coupling rings, it still brings $\frac{1}{2}$ phase-locking constraints.
Thus in this work, $\gamma$ and $\delta$ are set as -0.6 and -0.1 respectively.
The biological assumption for these parameter selections is that the top layer of hip joints dominates the rhythm of the whole network, and the bottom layer ``follows'' the rhythm generated by the top layer. Set $\gamma \neq \delta$ potentially breaks the global $\mathbf{D}_4$ symmetry, since it leads to inequivalent in a pair of hip and knee neurons, but the following numerical simulation proves this inequivalent does not affect gait control.

So far the eight-neuron network has been constructed, and we consistently utilized this network in subsequent simulations. Table~\ref{T:network parameter} lists the parameters of the network.

\begin{table}
\small\sf\centering
\caption{Parameters of the eight-neuron network}
\label{T:network parameter}
    \begin{tabular}{cccc}
        \toprule
         ~& Type&Parameter&Value or function\\
        \midrule
        \multirow{7}{*}{Neuron}&\multirow{4}{*}{variable}&$a(a^h,a^k)$&\multirow{4}{*}{gait control}\\
        &~&$f(f^h,f^k)$\\
        &~&$k_1(k_1^h,k_1^k)$\\
        &~&$k_2(k_2^h,k_2^k)$\\
       \cmidrule(lr){2-4}
       &\multirow{3}{*}{constant}&$b$&$-2000$\\
       &~&$p$&$10$\\
       &~&$q$&$30$\\
       \cmidrule(lr){1-4}
       \multirow{4}{*}{Network}&\multirow{4}{*}{constant}&$\alpha$&$-0.15$\\
       &&$\beta$&$-0.15$\\
       &&$\gamma$&$-0.6$\\
       &&$\delta$&$-0.1$\\
       \bottomrule
    \end{tabular}
\end{table}

\section{Numerical simulations}
\label{sec:Numerical simulations}
To demonstrate that the eight-neuron network model can effectively generate rhythmic gait signals and achieve gait transitions, we presented a series of numerical simulations.
The numerical simulations are carried out in Python 3.11, and the ODEs of the network are calculated by the Forth Runge-Kutta method.
The initial state of the eight-neuron network is listed in Table~\ref{T:initial state}.

\subsection{Five gaits}
\begin{figure*}[htbp]
  \centering
\begin{minipage}{1\linewidth}
  \centering
  \includegraphics[width=16cm]{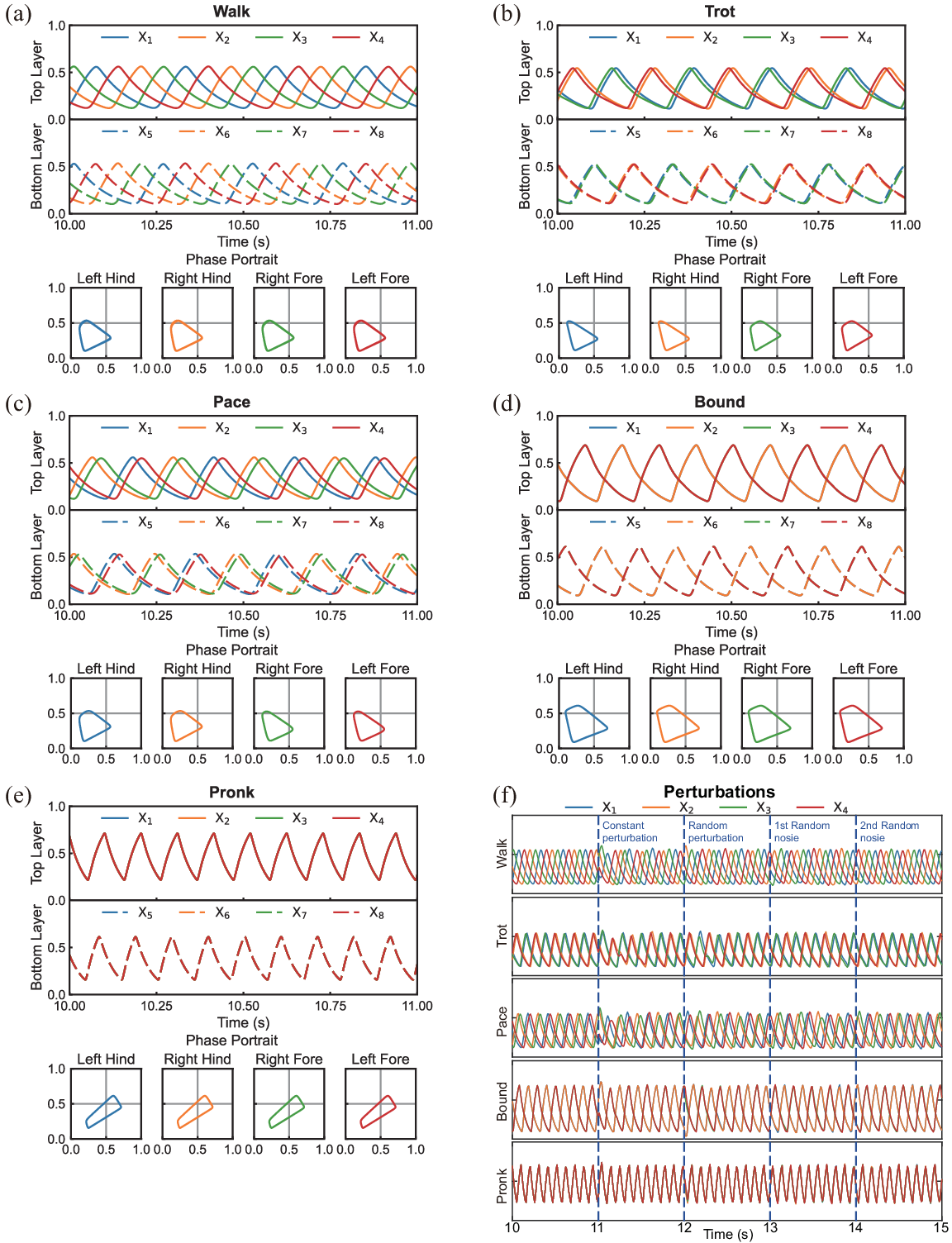}
  \caption{
  (a)-(e) Signals of the eight-neuron network corresponding to the walk, trot, pace, bound, and pronk gaits, respectively.
  (f) Tests of five gaits against four types of perturbations.
  At 11s, a constant perturbation 0.1.
  At 12s, a random perturbation within $[-0.08,0.08]$.
  In 13-14s, a random noise within $[-0.008,0.008]$.
  In 14-15s, a random noise within $[-0.005,0.005]$.
  }
  \label{fig:5gaits and perturbations}
\end{minipage}
\end{figure*}

The eight-neuron network is capable of generating five gaits: walk, trot, pace, bound, and pronk. The spatiotemporal symmetries of these gaits are shown in Fig.~\ref{fig:gaits_illustration}. The numerical simulation results of the gaits are shown in Fig~\ref{fig:5gaits and perturbations}. For each gait, the control parameters in simulation and the periods are listed in Table~\ref{T:gait parameter}. The phase portraits of the hip-knee neuron pairs are also illustrated to prove that the eight-neuron network can generate stable phase-locking between the hip and knee neurons. This phase-locking feature is helpful in the quadruped locomotion control for it can replace the mapping function by directly sending the signals of the bottom layer to the corresponding knee joints. The video of five gaits is provided in the \textcolor[rgb]{1,0,0}{video E1}.

Compared with the Stein model four-neuron $\mathbf{Z}_4$ network in ~\cite{collins_hardwired_1994}, it's found that the value of $a$ increased from the walk to the trot to the bound. However, pace gait is found to be not achievable in the four-neuron $\mathbf{Z}_4$ network.
In the numerical simulation, we observed the conjugate phenomenon of these two gaits, and we have studied in detail the rules involved in the transition between these two gaits.
Therefore, although the parameters we set for the trot and pace are very close, we are still able to achieve a successful gait transition (see next subsection).
However, we believe that in the eight-neuron network, there must be better combinations of control parameters for either trot and pace or other gaits.

Both trot and pace gaits have been observed in animals. Experiments also proved that some quadruped animals can learn pace gait by training~~\citep{blaszczyk_alteration_1989}, which implies that the CPG of the animals should support the rhythms of both trot and pace. From this aspect, the eight-neuron network which supports both trot and pace gaits is close to biology. Besides, assigned neurons to each joint with inner stability from the local symmetry would make the eight-neuron network in line with the biological characteristics of locomotion.

\begin{table}
\small\sf\centering
\caption{Initial state of the network}
\label{T:initial state}
    \begin{tabular}{c|ll}
        \toprule
        Neuron& \multicolumn{2}{c}{$(x_i,y_i,z_i)$}\\
        \cmidrule{1-3}
        1 and 5& $(1,0.04,0.016)$ & $(1,0.045,0.018)$ \\

        2 and 6& $(1,0.045,0.018)$ & $(0.8,0.05,0.02)$ \\

        3 and 7& $(0.8,0.05,0.02)$ & $(1,0.025,0.014)$ \\

        4 and 8& $(1,0.025,0.014)$ & $(1,0.04,0.016)$ \\

        \bottomrule
    \end{tabular}
\end{table}

\begin{table}
\small\sf\centering
\caption{Control parameters and periods of five gaits}
\label{T:gait parameter}
    \begin{tabular}{ccccccc}
        \toprule
          \multicolumn{2}{c}{\multirow{2}{*}{Variables}} &\multicolumn{5}{c}{Gait type}\\
         \cmidrule{3-7}
         ~&~&Walk&Trot&Pace&Bound&Pronk\\
        \midrule
        \multirow{4}{*}{\makecell{Top\\Layer}}& $a^h$ & $10$ & $11$ & $11$ & $16$ & $22$ \\
        ~& $f^h$ & $40$ & $41$ & $41$ & $50$ & $65$\\
        ~& $k_1^h$ & $0$ & $0.085$ & $0.04$ & $0.1$ & $0.3$ \\
        ~& $k_2^h$ & $0$ & $56$ & $54$ & $59$ & $60$ \\
        \cmidrule{1-7}
        \multirow{4}{*}{\makecell{Bottom\\Layer}}& $a^k$ & $10$ & $11$ & $11$ & $14$ & $22$ \\
        ~& $f^k$ & $40$ & $41$ & $41$ & $45$ & $65$ \\
        ~& $k_1^k$ & $0$ & $0$ & $0.01$ & $0$ & $0.2$ \\
        ~& $k_2^k$ & $0$ & $0$ & $54$ & $0$ & $60$ \\
        \midrule
        \multicolumn{2}{c}{Period (s)}&$0.259$&$0.224$ &$0.233$ &$0.213$ &$0.209$\\

        \bottomrule
    \end{tabular}

\end{table}

We further validated the stability of the gait by introducing perturbations, the stability features endow the CPG to be available for sensory feedback integration.
We assume that the bottom layer follows the hip layer, thus we applied perturbations only on top-layer neurons.
For each gait, we applied four types of perturbations:
a constant perturbation,
a random perturbation, and two types of random noises. The signals of the neurons are shown in Fig.~\ref{fig:5gaits and perturbations} (f).
Its shown that the walk and pace gaits are more affected by perturbations, but all five gaits can maintain the phase relations among legs against all perturbations.

\subsection{Gait transition strategy}
Gait transition endows the animals to switch their speed, terrain adaptability, and energy consumption~~\citep{alexander_principles_2003,xi_selecting_2016}.
The most attractive feature of CPG for gait generation is that the joint signals are continuous during gait transitions.
From the aspects of dynamical systems, the gait transition of the CPG refers to the bifurcation of the system~\citep{schoner_dynamic_1988,collins_symmetrybreaking_1992,collins_coupled_1993,golubitsky_singularities_2012}.
The change of the control parameters of the ODEs (in this work, the control parameters are $a, f,k_1$, and $k_2$) leads to the system switch from one attractor to the other.

In the eight-neuron network with five gaits, there are a total of twenty types of gait transitions.
It's worth mentioning that the phase relation between a pair of knee and hip neurons is always maintained, even in situations of transition failure.
Therefore, in the following discussion of transitions, our main focus will be on neurons 1-4 and not go into detail about neurons 5-8.

We applied four strategies to achieve all continuous transitions, which are \emph{Switch}, \emph{Power Pair}, \emph{Wait \& Switch}, and \emph{Wait \& Power Pair}. These strategies are demonstrated below:

\begin{itemize}
  \item[\textbf{S.1}] \textbf{\emph{Switch}}: Directly change the control parameters ($a, f,k_1$, and $k_2$) to the target gait. This is the simplest strategy, the applicable transitions are walk-to-\{bound, pronk\},\{trot,pace\}-to-\{walk, bound, pronk\}, and bound-to-pronk

  \item[\textbf{S.2}] \textbf{\emph{Power Pair}}: In addition to changing control parameters, the amplitude parameters of the driving signals for selected neurons are increased for brief periods and then return to their original values. This strategy has been previously proposed in ~\cite{collins_hardwired_1994}. In their four-neuron network, for transitions bound-to-walk and bound-to-trot, the selected neurons are a pair of diagonal neurons 1 and 3.
      In this work, the \emph{Power Pair} is applicable for transitions pronk-to-\{walk, bound\}.
      For these two transitions, adopting the \emph{Power Pair} at any moment can lead to success.

\end{itemize}

Increasing the driving signal can be considered as stimulating these neurons.
In our strategy, the shape of the driving signal is defined by four parameters: gain ratio $R_P$, duration period $T_P$, and duty cycles of rising edge and falling edge, $\eta_R$ and $\eta_P$, respectively.
The method of constructing the stimulation is provided in the Appendix C.

\begin{figure*}[htbp]

  \centering
  \includegraphics[width=17.5cm]{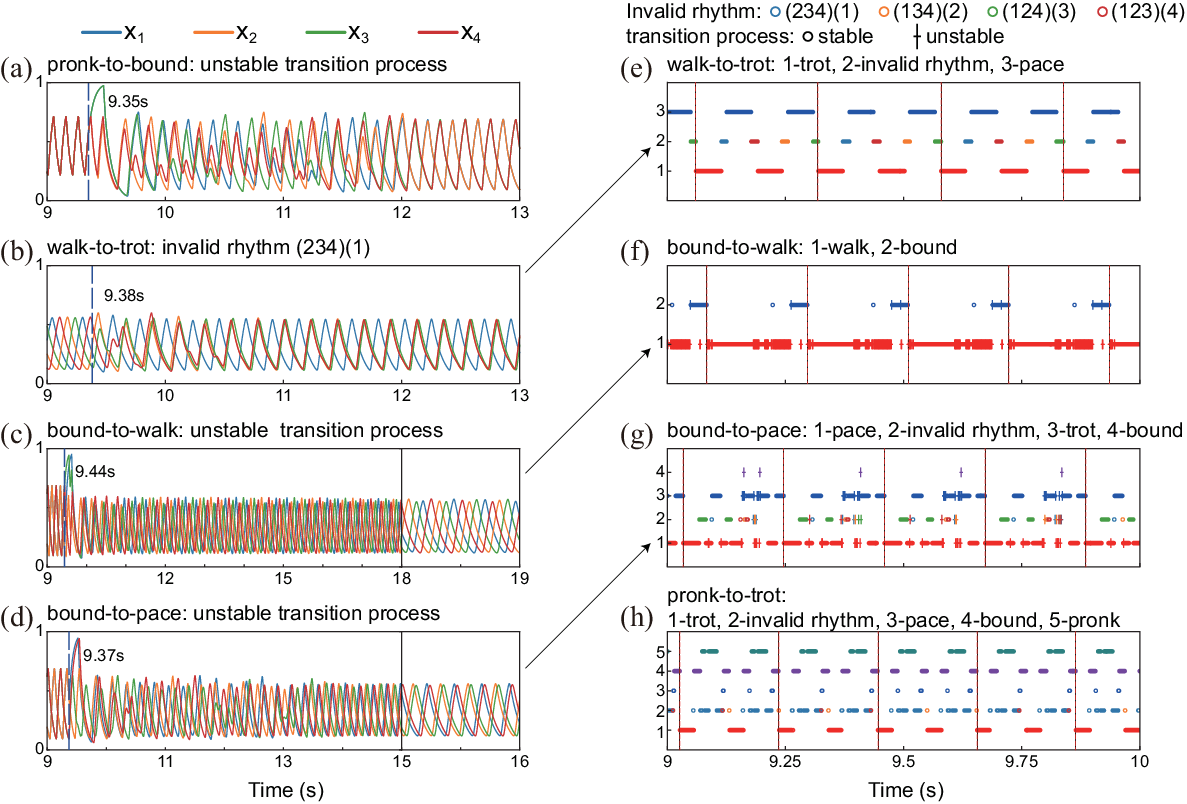}
  \caption{
  (a) Unstable transition process in pronk-to-bound under \emph{Power Pair} strategy with stimulated neurons 1 and 3. To solve this problem, stimulated neurons are selected as 1 and 2.
  (b) Invalid rhythm (234)(1) in walk-to-trot with \emph{Switch} strategy. To solve this problem, \emph{Wait \& Switch} strategy is proposed.
  (c) Unstable transition process in bound-to-walk with \emph{Power Pair} strategy. To solve this problem, \emph{Wait \& Power Pair} strategy is proposed.
  (d) Unstable transition process in bound-to-walk with \emph{Power Pair} strategy. \emph{Wait \& Power Pair} strategy is also applied here to avoid instability, invalid rhythm, and wrong gait.
  (e)-(h) Some selected patterns of transitions with difficulties under \emph{Switch} and \emph{Power Pair}.
  }
  \label{fig:difficulties-in-transitions}

\end{figure*}

During the pronk-to-bound transition, stimulating neurons 1 and 3 as in the pronk-to-walk transition would cause signal instability (Fig.~\ref{fig:difficulties-in-transitions}(a)).
At this condition, the amplitude of the neuron signals continues to fluctuate, and this state requires a long time (several seconds) to stabilize the walk gait.
On the other hand, we found that if excitatory neurons 1 and 2 are selected, the gait transition can avoid such fluctuations.
It can be observed from the waveform that, as neurons 1 and 2 are excited, their amplitudes increase and eventually enter the same phase, thus causing the rhythm to change to the bound gait, denoted $(12)(34)$ which refers to the rhythm that neurons 1 and 2 in phase, 3 and 4 in phase, $(12)$ and $(34)$ out of phase.
Therefore, we considered that when using the \emph{Power Pair} strategy, the choice of stimulated neurons should be those in the same phase in the target gait (trot-$(13)(24)$, pace-$(12)(34)$), and we maintained this approach in the strategies for \{bound, pronk\}-to-\{trot, pace\}.

Up to this point, we have found that the aforementioned strategies cannot be successful in a group transition: \{all gaits\}-to-\{trot, pace\} and bound-to-walk. These transitions can further be classified into Category \#1: \{walk, trot, pace\}-to-\{trot, pace\},  Category \#2: bound-to-walk and Category \#3: \{bound, pronk\}-to-\{trot, pace\}.

For Category \#1, the difficulty arises from the conjugation of the trot and pace. It is observed that the gait after transitions has several failed situations.
Taking the walk-to-trot transition as an example, we observed several types of failures after the switch, such as entering an unwanted target gait such as pace or entering an invalid rhythm, for instance, neurons 1, 2, 3 in phase, neuron 4 out of phase, which is denoted as $(123)(4)$ (Fig.~\ref{fig:difficulties-in-transitions}(b)).

We have found that the system state after applying the \emph{Switch} and the timing of applying the strategy are related, and these states exhibit periodicity.
It is shown in Fig.~\ref{fig:difficulties-in-transitions}(e) that, as the transition time increases, the system continually switches between valid gaits and invalid rhythms. This pattern can be summarized as ``trot $\rightarrow$ invalid rhythm $\rightarrow$ pace $\rightarrow$ invalid rhythm $\rightarrow$ trot $\rightarrow$ invalid rhythm $\rightarrow$  pace $\rightarrow$ invalid rhythm''.
To overcome the potential gait transition failures caused by this conjugation and the invalid rhythms, we proposed the third strategy, \emph{Wait \& Switch}.

\begin{itemize}
  \item[\textbf{S.3}] \textbf{\emph{Wait \& Switch}}: This is an improved strategy based on the \emph{Switch} strategy.
  Upon receiving the transition command, the \emph{Switch} is not executed immediately;
  instead, it is determined whether the current switch will result in the desired gait.
  If it would lead to an unwanted gait or invalid rhythm, the \emph{Switch} is delayed until the next appropriate opportunity.
  The criteria for the applicability of \emph{Wait \& Switch} are provided in the Appendix C.
  This strategy introduces a certain delay, which will not exceed a gait cycle, but it ensures the success of the gait transition.
\end{itemize}

\begin{figure*}[htbp]
  \centering
  \begin{minipage}{1\linewidth}
  \includegraphics[width=17cm]{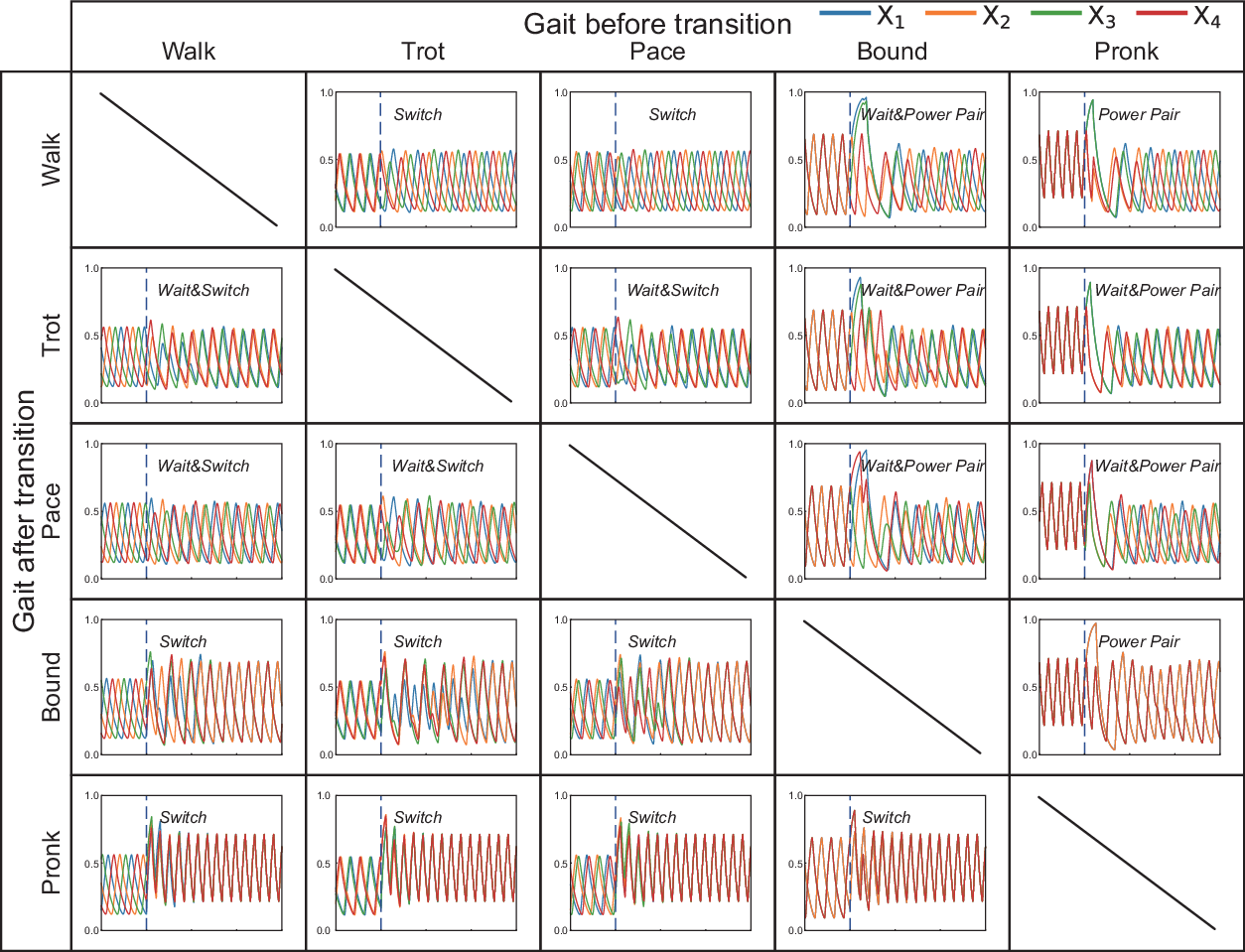}
  \caption{
  Twenty gait transitions with corresponding strategies \emph{Switch}, \emph{Wait \& Switch}, \emph{Power Pair} and \emph{Wait \& Power Pair}.
  Each transition spans two seconds. The dashed line marks the time of the transition execution.
  }
  \label{fig:transitions}
  \end{minipage}
\end{figure*}

\begin{figure}[t]
  \centering
  \includegraphics[width = 8cm]{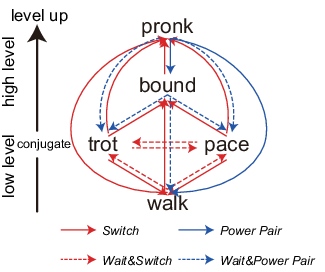}
  \caption{The strength levels of each gait and the corresponding gait transition strategies.
  }
  \label{fig:gait transition strategy}
\end{figure}

For the transitions in  Category \#2 and \#3, using either the \emph{Switch} or the \emph{Wait \& Switch} strategy does not induce a change in rhythmicity. Direct use of the \emph{Power Pair} can also lead to unwanted gaits or incorrect rhythms, thus we propose the last gait transition strategy, \emph{Wait \& Power Pair}.

\begin{itemize}
  \item[\textbf{S.4}] \textbf{\emph{Wait \& Power Pair}}: This is an improved strategy based on the \emph{Power Pair}.
  The method is similar to \emph{wait and switch}.
  It does not execute immediately upon receiving the transition command but waits for the right moment to make the \emph{Power Pair}.
  The criteria for the applicability of the strategy are provided in the Appendix C.
\end{itemize}

For Category \#2: bound-to-walk, ~\cite{collins_hardwired_1994} have previously reported that in the four-neuron network, using \emph{Power Pair} to stimulate the ipsilateral neurons 3 and 4 for implementing bound-to-walk transitions carries a probability of failure.
In our network, we have found that even though the selection are diagonal neurons 1 and 3, \emph{Power Pair} still leads to several incorrect conditions

A phenomenon that had not been observed in previous transitions is that the neurons would enter a chaotic state that lasts for an extended period, but eventually, it would return to a stable walking gait, which we refer to as an unstable transition process, which is shown in Fig.~\ref{fig:difficulties-in-transitions}(c).
The signal after \emph{Power Pair} can be categorized as: walk gait (stable and unstable) and bound gait and unstable transition process, and the pattern of the transition is shown in Fig.~\ref{fig:difficulties-in-transitions}(f).

Although the unstable signals display the rhythmicity of walk gaits in the short term, the amplitude of the signal from each neuron is not stable, showing periodic variations. It is observed that, over a longer period, the disordered signals will gradually return to a stable walk gait. Considering this period may lead to the failure of the locomotion, we adopted the \emph{Wait \& Power Pair} to ensure the success of the transition.

For Category \#3: \{bound, pronk\}-to-\{trot, pace\}, unstable transition process and invalid rhythms caused by conjunction are both observed.
Taking bound-to-pace as an example, the unstable neuron signals are shown in Fig.~\ref{fig:difficulties-in-transitions}(d), and the patterns are shown in Fig.~\ref{fig:difficulties-in-transitions}(g).
The patterns of pronk-to-trot are shown in Fig.~\ref{fig:difficulties-in-transitions}(h), it's observed that the rhythms of the neuron signal after \emph{Power Pair} are very complex. Many conditions may lead to the failure of the gait transition, which further proves the necessity of implementing gait transitions at appropriate moments by utilizing \emph{Wait \& Power Pair}.

The simulation results of the twenty gait transitions are demonstrated in Fig~\ref{fig:transitions}.
The simulations are provided in the \textcolor[rgb]{1,0,0}{video E2}.
The proposed four strategies ensure the success of transitions among all types of gaits and the continuity of neuronal signals.

\subsection{Brief discussion}
We can rank the five gaits according to the ``strength'' of their corresponding attractors and summarize the applicable situations for gait transition strategies (Fig.~\ref{fig:gait transition strategy}). Low level gaits include walk, trot, and pace, high level gaits are bound and pronk. Based on this classification, we can summarize two regularities:
\begin{itemize}
  \item[\textbf{R.1}]\emph{Switch} and \emph{Wait \& Switch} are applicable to: \{low level gaits\}-to-\{all gaits\}, \{each gait\}-to-\{gaits higher than itself\}.
  \item[\textbf{R.2}]\emph{Power Pair} and \emph{Wait\&Pair Power} are applicable to: \{high level gaits\}-to-\{gaits lower than itself\}.
\end{itemize}
Another regularity can be summarized from the conjunction property of the trot and pace, which is :
\begin{itemize}
  \item[\textbf{R.3}]\emph{Wait \& Switch} and \emph{Wait \& Power Pair} are applicable to: \{all gaits\}-to-\{trot, pace\}.
\end{itemize}
The only exception is the transition bound-to-walk. This transition also requires the use of a \emph{Wait \& Power Pair} strategy because unwanted bound gait and unstable transition process may occur.
The following research will be focused on how to optimize the strategy parameters or the selection of stimulation to improve this phenomenon.

It is worth mentioning that, gait transitions can not only be achieved by changing neuron dynamic parameters from the MLR but also by altering the coupling parameters of the network. In this study, we do not employ the latter method because changing network coupling parameters is more suitable for the couplings with linearized form~\citep{buono_models_2001,rutishauser_passive_2008,song_gaits_2023}. In the Stein model, we have integrated coupling into neuron dynamics, manifesting in a nonlinear form, as in equations~(\ref{eq:stein model}) and (\ref{eq:driving signals}).
In this case, it becomes difficult to verify the eigenvalues of the coupling matrix, and using the coupling matrix to regulate bifurcation is also challenging.

In summary, we verified through numerical simulations that using the Stein neuronal model as an eight-neuron network can generate five types of gaits, and we verified the stability of the gaits by applying perturbations.
For twenty types of gait transitions, we proposed four strategies of gait transition based on their types. We discussed in detail the design rationale, executing methods, and applicable conditions of these transition methods.
All four gait transition methods involve changing parameters regulated by the MLR, thus ensuring that the signals generated by the neurons during all gait transitions are continuous.

\begin{figure*}[ht]
  \centering
  \includegraphics[width = 17.5cm]{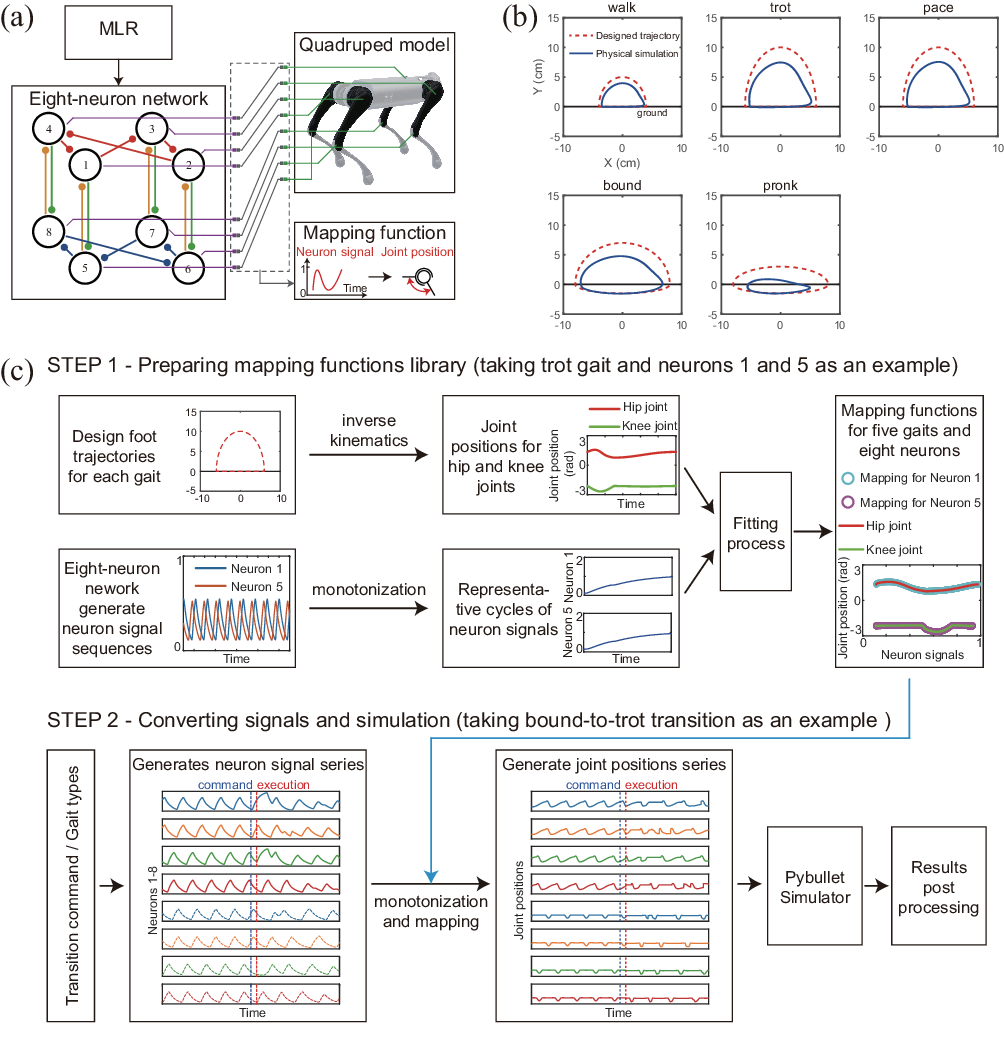}
  \caption{
  (a) Neuron assignments and signals processing of the physical simulation. Totally eight hip and knee joints have assigned neurons. The neuron signals are dimensionless quantities. The mapping functions library is introduced to convert neuron signals into joint positions.
  (b) Foot trajectories corresponding to five gaits. Each foot trajectory has the hip joint coordinates at (0, 20).
  (c) Two step framework of the physical simulation. The first step is to prepare a mapping function library using neuron signals and foot trajectories. The second step is to use the mapping function library to convert neuronal signals into joint position signals and to perform the simulation.
  }
  \label{fig:control_architecture}
\end{figure*}

\section{Physical simulation}

To demonstrate how to use the eight-neuron network for gait control of a quadruped robot and to prove the stability of the network during gait transitions, we presented a series of physical simulations. We used PyBullet 3.2.6~\citep{coumans_pybullet_2024} as the simulator with Unitree Robot Go1~\citep{unitree_unitree_2024a,unitree_unitree_2024} as the robot model.
In the physical simulation, the hip-knee joint refers to the hip flexion-extension and knee flexion-extension DoF of the robot.
The abduction/adduction DoF of the hip joint is locked.

\subsection{Simulation framework}

Following the last section, the eight-neuron network is modeled in Python. The parameters that control the gaits (Table~\ref{T:gait parameter}) are considered as the MLR modules. After the gait parameters are set in MLR, the eight-neuron network generates the signals, and each neuron is assigned to a joint of the robot, and the neuron signals are transferred into the joint position signals through the mapping functions. (Fig.\ref{fig:control_architecture}(a)). Before conducting the physical simulation, we first pre-design a foot trajectory for each of the five gaits (Fig.\ref{fig:control_architecture}(b)).

The physical simulation framework can be divided into two steps (Fig.\ref{fig:control_architecture}(c)).
The first step is the preparation, where mapping equations are generated for each neuron and each gait foot trajectory, forming a library of mapping functions.
Here is a demonstration, using the trot gait as an example.
For the designed foot trajectory, inverse kinematics is adopted for generating the joint positions for a foot trajectory cycle.
On the other hand, neurons 1 and 5 are selected as a pair, and their signals for trot gait are processed. These signals are monotonized into continuously increasing signals within each period, and one period is taken from each of the two neurons.
The joint position signals and the monotonized neuron signals are fitted to create mapping functions. Neurons 1-4 are inversely synthesized to the hip joints, and neurons 5-8 are fitted to the knee joints. This process is executed for all gaits and all neurons to establish a library of mapping functions.

The second step is generating the corresponding neuron signal sequences for the simulation, which could be a single gait or a sequence of gait transitions. By using the aforementioned library of mapping functions, the neuron signals corresponding to the gait and neurons are transformed into joint angle signals, which are then sent to the PyBullet simulator for execution.
Here is a demonstration, using the bound-to-trot transition as an example.
First, the MLR generates the signal sequences for eight neurons based on the gait type and transition time, using an appropriate switching strategy. Since the bound-to-trot transition uses the \emph{Wait \& Power Pair}, there is a certain delay between the execution time and the issuance time of the transition command.
The neuron signal sequences are further monotonized and converted into joint position signals through corresponding mapping functions.
These signals correspond to the joints of the robot model and are sent to the simulator.
It can be observed that during the gait transition process, the signals of all joints remain excessively smooth.
This ensures a high success rate of gait transition from the perspective of joint signals.

\begin{figure*}[ht]
  \centering
  \includegraphics[width=17cm]{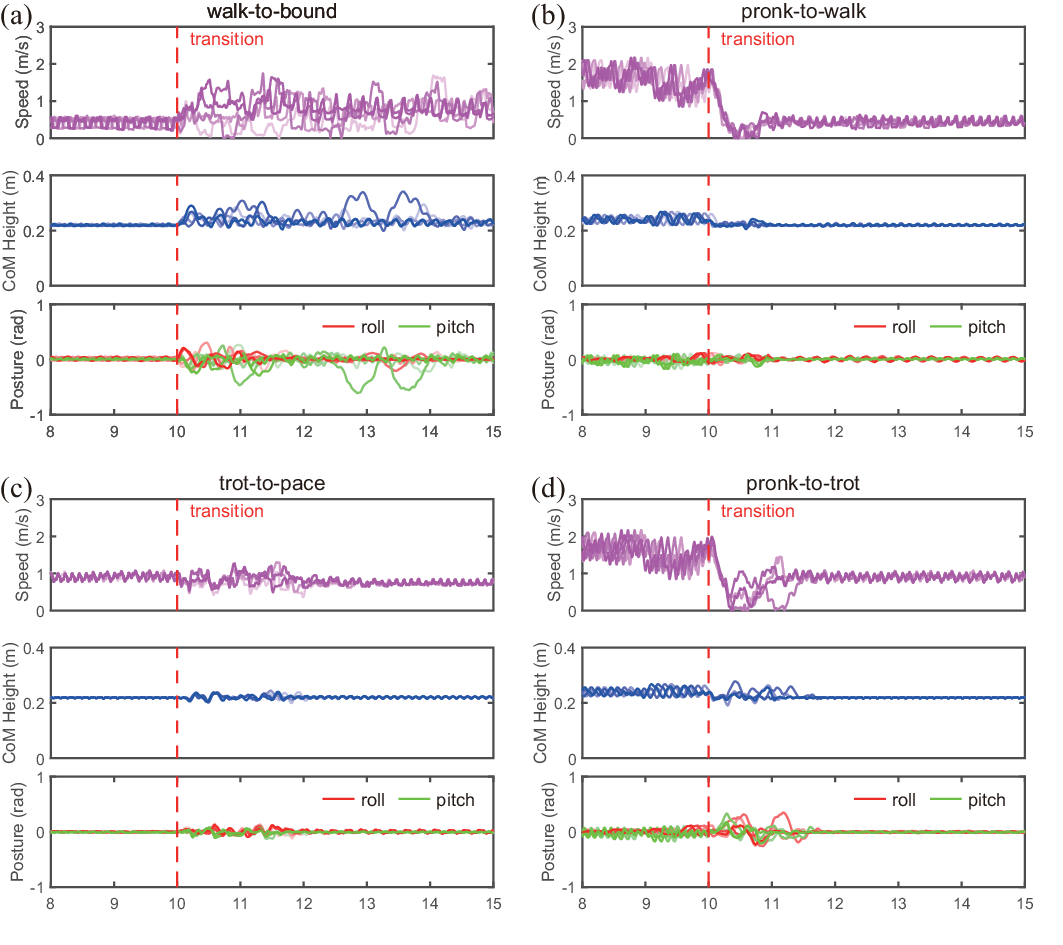}
  \caption{
  The posture, speed, and center of mass height during the four gait transitions. Five trials are reported for each transition.
  All data is aligned according to the actual execution time of the transition.
  (a) walk-to-bound with \emph{Switch} strategy.
  (b) pronk-to-walk with \emph{Power Pair} strategy.
  (c) trot-to-pace with \emph{Wait \& Switch} strategy.
  (d) pronk-to-trot with \emph{Wait \& Power Pair} strategy.
  }
  \label{fig:gait-transition-physical-simulation}
\end{figure*}

The simulation animations for five gaits are provided in the \textcolor[rgb]{1,0,0}{video E3}. We calculated the model's speed after the neuron signals of the eight-neuron network stabilized, these data are listed in Table~\ref{T:Speed_Simulation}.
Due to the impact-rich nature, the speed of the bound gait is not stable. One potential reason is that the quadruped robot model we are using does not include the elastic element, In previous studies, the elastic element has been considered an important factor in reducing impact during the touch-down process~\citep{remy_stability_2010, hyun_high_2014}.
Except for the speed fluctuations in the bound gait, all other gaits in the simulation exhibited stability.
These results indicate that the eight-neuron network, combined with a simple simulation framework, can achieve stable motion control.

\subsection{Physical simulation of gait transition}

\begin{table}
\small\sf\centering
\caption{Physical Simulation Performance}
\label{T:Speed_Simulation}
    \begin{tabular}{ccc}
        \toprule
        {Gait type}&{Speed (m/s)}&{$\text{STD}_\text{speed}$ (m/s)}\\
        \midrule
        Walk&0.42&0.11\\
        \cmidrule(lr){1-3}
        Trot&0.91&0.07\\
        \cmidrule(lr){1-3}
        Pace&0.75&0.05\\
        \cmidrule(lr){1-3}
        Bound&0.82&0.42\\
        \cmidrule(lr){1-3}
        Pronk&1.56&0.27\\
        \bottomrule
    \end{tabular}
\end{table}

\begin{table}
\footnotesize\sf\centering
\caption{Success rate of twenty gait transitions in physical simulation.}
\label{T:success rate}
    \begin{tabular}{ccccccc}
        \toprule
        && \multicolumn{5}{c}{Gait before transition}\\
        \cmidrule{3-7}
        && \rotatebox[origin=c]{00}{Walk} & \rotatebox[origin=c]{00}{Trot} & \rotatebox[origin=c]{00}{Pace} & \rotatebox[origin=c]{00}{Bound} &\rotatebox[origin=c]{00}{Pronk}\\
        \cmidrule{1-7}
        \multirow{5}{*}{\rotatebox[origin=c]{90}{\makecell*[c]{Gait after\\ transition}}} &\rotatebox[origin=c]{00}{Walk}& \textbackslash &98.8\% &96.2\% &98\%&99\% \\

        &\rotatebox[origin=c]{00}{Trot}& 100\% & \textbackslash&100\% &100\%&99.8\% \\

        &\rotatebox[origin=c]{00}{Pace}& 100\% &100\%  &\textbackslash &100\%&100\% \\

        &\rotatebox[origin=c]{00}{Bound}& 97\% &86.8\% &90\% &\textbackslash &71.8\%\\

        &\rotatebox[origin=c]{00}{Pronk}& 100\% &98.6\% &96\%&92.2\%&\textbackslash\\
        \bottomrule
    \end{tabular}
\end{table}

Based on the above simulation framework, we investigated the performance of the eight-unit network during gait transitions.
We first tested the success rate of twenty gait transitions.
For each type of gait transition, we tested 500 transitions during the 10-11 second period at intervals of 0.02 seconds, covering at least four cycle lengths for each transition to ensure the generality of the results.
The success rates are listed in Table~\ref{T:success rate}.
It can be observed that, except for a few transitions related to bound and pronk, the success rate of most transitions remains above 90\%, with some transitions achieving a 100\% success rate.
These results indicate that the eight-neuron network can achieve stable motion control and gait transition through a simple signal mapping method.

We further investigated several representative processes of gait transitions in detail.
The selected transition and the corresponding metrics as speed, center of mass height and posture are shown in Fig.~\ref{fig:gait-transition-physical-simulation}, and the animation of simulations is provided in \textcolor[rgb]{1,0,0}{video E4}.
The results imply that when a transition happens from a higher level gait to a lower level gait, the robot may experience excessive deceleration, but it eventually returns to a normal level.
During the transitions, the robot's center of mass height and posture will experience some disturbances, but they will eventually return to normal levels.

\subsection{Top layer for foot trajectory control}
In previous studies, the classic application scenario of CPG implanted as a gait generator is responsible only for controlling the phase relationships between the four legs, without specifically controlling each joint within a leg. The top layer in our proposed eight-neuron network, which is used to control all hip joints, is sufficient for this scenario.

\begin{figure}[t]
  \centering
  \includegraphics[width = 8cm]{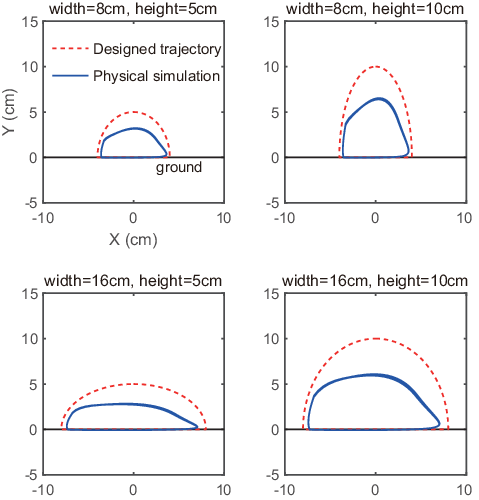}
  \caption{
  Four foot trajectories with walk gait achieved by top layer as gait generator through inverse kinematics.
  }
  \label{fig:single-layer-physical-simulation}
\end{figure}

This approach, while maintaining the diversity of gaits and the continuity of joint signals inherent to the network, allows for the convenient design of the robot's foot trajectory through inverse kinematics. Here, we demonstrate this approach in a walk gait. We design four different trajectories (Fig.~\ref{fig:single-layer-physical-simulation}), and only the top-layer neurons are involved in phase relation control, with the joint positions calculated by inverse kinematics. The animations of these trials are provided in \textcolor[rgb]{1,0,0}{video E5}.

\subsection{Brief discussion}
In this section, We have demonstrated through physical simulations that gait control of a commercial robot model can be achieved using a simple framework. Even without any real-time feedback, the robot exhibited a high level of stability and success rate.
In addition, the eight-neuron network can also, just like other classic CPGs, control the phase among legs using only its top-layer network. Moreover, there is potential for these two approaches to be integrated in subsequent research to simultaneously achieve coordinated control of knee and hip joints and foot trajectory design.
These results confirm the potential of the proposed eight-neuron network for directly controlling quadruped robots, which inherently possess high stability, or as a gait generator integrated into existing control architectures.

\begin{figure}[t]
  \centering
  \includegraphics[width = 7.5cm]{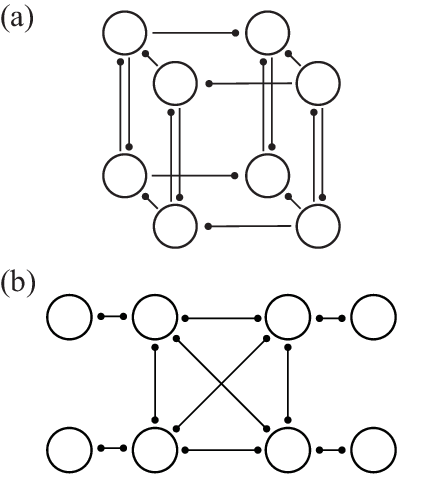}
  \caption{
  Two similar network architectures.
  (a) The $\mathbf{D}_4$ network proposed in~\cite{buono_models_2001}.
  (b) In our eight-neuron network, when $\beta = 0$, the bottom layer for knees is loose. At this condition, the relationship between hip and knee neurons is similar to~\cite{billard_biologically_2000}.
  }
  \label{fig:two-similar-networks}
\end{figure}

\section{Discussion}
\subsection{Comparison with Buono and Golubitsky's Work}
Martin Golubitsky and his collaborators conducted pioneering research on symmetry and coupled-cell networks, which has provided theoretical guidance for the design of CPGs in robotics.  A type of eight-neuron network with $\mathbf{D}_4$ symmetry was previously introduced by \cite{buono_models_2001}. \cite{righetti_design_2006,righetti_pattern_2008} have successfully applied the $H/K$ theorem to design CPGs for several robots. The four-neuron network  with $\mathbf{D}_4$ symmetry has also been investigated by~\cite{righetti_control_2008}.

From the aspects of architecture, our eight-neuron network resembles Buono and Golubitsky's $\mathbf{D}_4$ network (Fig.~\ref{fig:two-similar-networks}(a)). Nonetheless, to control the hip and knee joints in quadruped locomotion, we designed our network by expanding from a four-neuron network. This involved replacing a single neuron with a pair of neurons and introducing local symmetry.
As a result, after constructing the network architecture, we further modified it to establish fixed neuron-to-joint assignments and regulated the phase relationships between the hip and knee joints by disrupting the local symmetry.
The unique features and modifications of our network are:
\begin{enumerate}
  \item {\it Group generator:}
  The eight-neuron network in our work can be considered as an expansion of the four-neuron $\mathbf{D}_4$ network.
  The $H/K$ theorem calculation for five gaits is based on the four-neuron network.
  To split the neuron into a two-neuron group, we introduced $\mathbf{Z}_2$ symmetry within a neuron group, resulting in a different group generator $\kappa$, distinguishing our methodology from the work of Buono et al.
  \item {\it Neuron assignment:}
  In Buono et al., both $\mathbf{Z}_4\times\mathbf{Z}_2$ network and $\mathbf{D}_4$ network have two layers. Signals generated by neurons in only one layer were assigned to the legs, while the function of the other layer serves to maintain signal propagation within the network~\citep{golubitsky_modular_1998} or to control the different muscle groups within a single leg~\citep{golubitsky_symmetry_1999}.
  The network of Buono et al. can control the phase among legs, but not the phase between hip and knee joints within a single leg. This is due to the strict adherence to complete symmetry between the upper and lower layers in their network, resulting in phase-locking within a bidirectionally connected pair of neurons. In our network, phase relation among legs is generated by the global symmetry, while the phase relation between hip and knee joints within a leg is regulated by the local symmetry, thus each neuron in our network is assigned to hip or knee joints, and the assignments are maintained.

  \item {\it Coupling and network modification:}
  Based on the above two points, in designing our network to control the knee and hip joints, we broke the $\mathbf{Z}_2$ symmetry through modification of the couplings ($\beta \neq \alpha$ resulted in the asymmetry of the unidirectional four-neuron rings between the upper and lower layers, $\gamma \neq \delta$ led to asymmetry in a pair of neurons). Therefore, in our network, the group generator $\omega$ was maintained, but both $\kappa$ and $\lambda$ were invalidated.
  Since $\lambda$ has been broken, it allows us to regulate the phase relation between a pair of neurons, thereby achieving control of the foot trajectory based on the position of the knee and hip joints.
\end{enumerate}

\subsection{Assumption: knee network follows the hip network}
\label{sec:following_effect}
In section \ref{sec:Design couplings}, we propose the assumption that the knee network follows the hip network.
From the aspect of the neural model, in Table~\ref{T:gait parameter}, the values of the gait control parameters $a$, $f$, $k_1$, and $k_2$ in the top layer are commonly larger than in the bottom layer. $a^k \geq a^h$ implies that the neurons for hip joints are more active than the neurons for knee joints. $f^h\geq f^k$, $k_1^h\geq k_1^k$, $k_2^h\geq k_2^k$ imply that the driving signals the neurons received from the MLR are different, where the signals for neurons in the top layer are stronger than the neurons in the bottom layer.

From the aspect of network architecture, the assumption leads to $\gamma \neq \delta$. $\gamma = -0.6$, $\delta = -0.1$ refers to a strong inhabitation from the top to the bottom and a weak inhabitation from the bottom to the top, respectively.
A more radical modification is to tune the value of the $\alpha$ and $\beta$. For instance, $\alpha=-0.15$, $\beta=-0.1$ refers to that the top layer is a strongly coupled network, and the bottom layer is a weakly coupled network. This will bring in flaws in the global symmetry of the whole network. However the numerical simulation of this network shows it can also perform five gaits and transitions successfully, but the control parameter combinations are different.

An exception is that $\beta = 0$, which means the bottom layer for knees is loose, and the network architecture is similar to \cite{billard_biologically_2000} (Fig.~\ref{fig:two-similar-networks}(b)). In our network, even though there are still eight neurons, the symmetry of the network descends to $\mathbf{Z}_4$. We attempted but did not find the pace gait. This result implies that the bottom layer is not only following the top layer but also affects the spatiotemporal symmetry of the whole network to generate the pace gait.

\subsection{Local symmetry: phase-locking and adaptive adjustment}
\label{sec:couplings_discussion}
An important advantage of the eight-neuron network we proposed is the adaptability and stable phase-locking between the hip and keen neurons. Attribute to the nonequivalent mutual inhibition $\gamma \neq \delta$, the phase relation between a pair of hip and knee neurons can be tuned by changing the value of $\gamma$ and $\delta$.

\begin{figure*}[t]
  \centering
  \includegraphics[width= 16 cm]{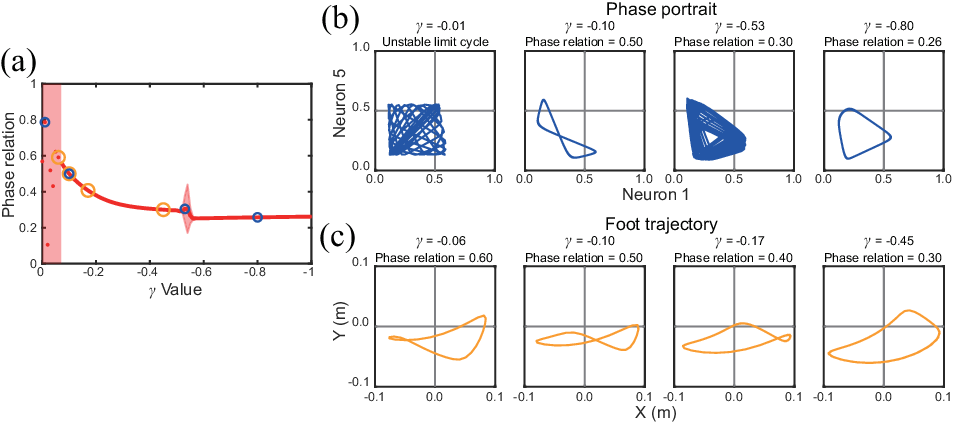}
  \caption{
  (a) The phase relation between neuron 1 and neuron 5 varies depending on the value of $\gamma$.
  (b) Four phase portraits of Neuron 1 and 5 with $\gamma$ selected as -0.01, -0.10, -0.53, and -0.80.
  (c) Foot trajectories achieved through forward kinematics with $\gamma$ selected as -0.06, -0.10, -0.17, and -0.45.
  From left to right, the foot trajectories gradually become more reasonable.
  }
  \label{fig:phase relation}
\end{figure*}

Here we give a brief example.
The tested network is the same as in the numerical simulations section, but only $\gamma$ is tuned, other parameters remain consistent with the settings of the walk gait.
Fig.~\ref{fig:phase relation} shows the phase relation between neurons 1 and 5 varies depending on the value of $\gamma$ ranges from 0 to -1. It shows that the majority of the values of $\gamma$ can generate a stable limit cycle, thus the phase relation is relatively stable. Some values of $\gamma$ (such as -0.01 and -0.53) can not generate a stable limit cycle, especially $\gamma=-0.01$, such parameters indicate a weak following of the bottom layer, which should be avoided during design.
The phase relation of the hip and knee neurons can be tuned ranging from 0.25 to 0.62 with a stable limit cycle.

When the phase relationships between neurons change with $\gamma$, the foot trajectory of the robot also changes accordingly. Here, we demonstrate the process of foot trajectory adjustment based on the following linear mapping to converter neuron signals to joint positions:
\begin{equation}\label{eq.mapping}
    \begin{aligned}
    \theta^{hip}_i&=2.2\times(x_i-0.38)+0.76, &i=1,2,3,4\\
    \theta^{knee}_i&=0.9\times(x_i-0.35)-2.08, &i=5,6,7,8\\
    \end{aligned}
\end{equation}

The foot trajectories are calculated through leg kinematics (Fig.~\ref{fig:phase relation}),
it shows that as the $\gamma$ parameter is adjusted, causing changes in the phase relationship between the knee and hip neurons, the foot trajectory gradually becomes more reasonable.
This characteristic provides an interface, making the eight-neuron network a promising research platform for embodied intelligence.
By integrating sensory information and adjusting the coupling strength of neurons, the robot's gait and foot trajectory can gradually adapt to the terrain.

\section{Conclusion}
In this work, we proposed an eight-neuron network as CPG for quadruped locomotion control. We utilize the $H/K$ quotient theory to design the symmetries of the network, enabling it to generate five gaits and control eight degrees of freedom of the knee and hip joints in a quadruped robot. We conducted detailed research on the gait transition of the network, explaining various conditions and patterns during gait transitions. Based on the discovered patterns, we designed several strategies to ensure the successful transition among all gaits. Furthermore, we investigated the potential applications of this network and demonstrated, through physical simulations, that by simply mapping signals, the network can be used for gait generation and motion control in quadruped robots.

CPGs are believed to widely exist in the neural systems of animals. In the field of robotics, various CPGs have been designed by drawing inspiration from the neural principles of biology for gait generation. In this work, we refer to observed phenomena and principles in biology and strive to design the architecture and strategies of neural networks from multiple perspectives, including biology, robotics, and computational neuroscience. For instance:
\begin{itemize}
  \item Based on the assumption of the knee joint following the hip joint, we designed the global and local symmetries of the neural network, as well as the foot trajectory modulation inspired by it.
  \item Based on the rhythmicity of gaits during transitions, we designed the stimulation selection method.
  \item Based on observed gaits in biology, we classified the conjugate characteristics and intensity levels of attractors corresponding to different gaits, and summarized the gait transition strategies employed.
\end{itemize}
We hope this research provides new tools for quadruped locomotion control and insights into biological CPG mechanisms.
From the perspective of quadruped locomotion control, we use the top layer of the eight-neuron network for hip control and the bottom layer for knee control. However, there are other possibilities for the signal assignments of each neuron. For example, a pair of neurons could be used to control a pair of antagonistic muscles. Considering the symmetry of muscle movement, the network proposed in this paper can be further expanded, or larger networks can be constructed to fit the biological motion mechanism, by using the method of neuron groups described before (Fig.~\ref{fig:8neuron}(a)).
We noticed that there is a primary gait jump that has not been found (Fig.~\ref{fig:gaits_illustration}). In this gait, the phase difference between the front and hind legs is 1/4, so we speculate that this gait may correspond to a generator similar to walk gait. This observation implies that the CPG in biology should be more complex than what we have proposed.

In conclusion, our proposed eight-neuron network serves as a new gait generator with multiple advantages, including high controllable degrees of freedom, diverse gait types, stable signals, continuous signal transitions during gait transition, and easy applicability. This not only provides a new gait generator for quadruped robots but also offers an example of the design and modulation of novel CPGs based on symmetry. Meanwhile, the designed network may also assist in decoding nervous systems and understanding biological control mechanisms.

Future work will focus on the following three topics.
\begin{itemize}
\item Expanding the structure of the network, such as controlling additional degrees of freedom in the hip joint, ankle joint, and waist joint.
\item Incorporating sensory networks in the form of reflexes to modulate gaits and foot trajectories.
\item Applying the eight-unit network in the motion control of the BioARS system ~\citep{liu_bioars_2020}.
\end{itemize}

BioARS is a project of constructing a small quadruped robot assembled from robotic insects~\citep{liu_s2worm_2022,liu_singularity_2024}. The simplicity and adaptability of CPG hold significant potential for the motion control of robots with onboard microcontrollers.
By leveraging the low computational power and distributed nature of the proposed eight-neuron network, we aim to achieve joint control for this insect-scale assembled quadruped robot.

\begin{acks}
This work was supported by the National Natural Science Foundation of China (Nos. 12321002, 91748209) and the 111 Project (No. B21034).
\end{acks}

\bibliographystyle{SageH}
\bibliography{reference_8neuronnetwork}

\clearpage

\section*{Appendix A. Index to multimedia extensions}
\begin{table}[h]
\footnotesize\sf\centering
\caption*{Table of multimedia extensions}
\label{T:media}
    \begin{tabular}{lll}
        \toprule
        Extension   &Media type &Description\\
        \midrule
        1   &Video  &\makecell[l]{Numerical simulation of five gaits.}\\
        \midrule
        2   &Video  &\makecell[l]{Numerical simulation of gait transitions\\ and demonstration of strategies.}\\
        \midrule
        3   &Video  &\makecell[l]{Physical simulation of five gaits.}\\
        \midrule
        4   &Video  &\makecell[l]{Physical simulation of gait transitions.}\\
        \midrule
        5   &Video  &\makecell[l]{Physical simulation of utilizing top layer\\ of the network as gait generator.}\\
        \midrule
        6   &Code  &\makecell[l]{Eight-neuron network with the physical\\ simulation.}\\
        \bottomrule
    \end{tabular}
\end{table}

\section*{Appendix B. A brief introduction of symmetry and group theory in CPG}
\label{App.Background of symmetry}
\textbf{Symmetry of network architecture:}

Taking the four-neuron ring in Fig.~\ref{fig:4neuron}(b) as an example,
the symmetry group of this loop is $\mathbf{Z}_4(\omega)$, $\omega=(1324)$.
$\mathbf{Z}_4$ is the cyclic group of order 4, $\omega$ is the group generator.
$\omega=(1324)$ refers to the permutation of the neurons.
Here the $\omega$ is written in cycle notation for simplification, and the complete form of a general permutation is denoted as :
\begin{equation}\label{eq:permutation operation general}
\sigma=\left(\begin{array}{ccccc}
x_1 & x_2 &  \cdots & x_n \\
\sigma\left(x_1\right) & \sigma\left(x_2\right) &  \cdots & \sigma\left(x_n\right)
\end{array}\right)
\end{equation}

The group operation is defined as:
\begin{equation}\label{eq:group operation}
\begin{aligned}
\sigma_1 \cdot \sigma_2 =&\left(\begin{array}{ccccc}
x_1 & x_2 &  \cdots & x_n \\
\sigma_1\left(x_1\right) & \sigma_1\left(x_2\right) &  \cdots & \sigma_1\left(x_n\right)
\end{array}\right)\cdot\\
&
\left(\begin{array}{ccccc}
x_1 & x_2 &  \cdots & x_n \\
\sigma_2\left(x_1\right) & \sigma_2\left(x_2\right) & \cdots & \sigma_2\left(x_n\right)
\end{array}\right)\\
 =&
\left(\begin{array}{ccccc}
x_1 & x_2 & \cdots & x_n \\
\sigma_1(\sigma_2\left(x_1\right)) & \sigma_1(\sigma_2\left(x_2\right))  & \cdots & \sigma_1(\sigma_2\left(x_n\right))
\end{array}\right)
\end{aligned}
\end{equation}

The complete form of the $\omega$ is
\begin{equation}\label{eq:permutation omega}
\omega=\left(\begin{array}{ccccc}
1 & 3 & 2 & 4 \\
3 & 2 & 4 & 1
\end{array}\right)
\end{equation}
which means the operation of $\omega$ is $\omega(1)=3$, $\omega(3)=2$, $\omega(2)=4$, $\omega(4)=1$.
The permutation of $\omega$ applied twice can be represented as:
\begin{equation}\label{eq:permutation omega twice}
\begin{aligned}
\omega &=\left(\begin{array}{ccccc}
1 & 3 & 2 & 4 \\
3 & 2 & 4 & 1
\end{array}\right)\cdot\left(\begin{array}{ccccc}
1 & 3 & 2 & 4 \\
3 & 2 & 4 & 1
\end{array}\right)\\
&=\left(\begin{array}{ccccc}
1 & 3 & 2 & 4 \\
2 & 4 & 1 & 3
\end{array}\right)
\end{aligned}
\end{equation}

This permutation preserves the architecture of the network (both the nodes and the coupling remain unchanged).
Symmetry group $\mathbf{Z}_4$ has four elements $(e\equiv\omega^4,\omega,\omega^2, \omega^3)$. $e$ is called identity element, after four times of $\omega$ permutation, the network return to its initial state.

Consider a counterexample, note that $\kappa=(13)(24)$ is not a group generator of this network.
The complete form of $\kappa$ is:
\begin{equation}\label{eq:permutation operation kappa}
\sigma=\left(\begin{array}{ccccc}
1 & 3 & 2 & 4 \\
3 & 1 & 4 & 2
\end{array}\right)
\end{equation}

The kappa permutation involves swapping neurons 1 and 3, as well as 2 and 4.
After the $\kappa$ permutation, the direction of the coupling ring becomes opposite to the original network.

\vspace{0.2cm}
\textbf{Symmetry of gait:}

Still taking Fig.~\ref{fig:4neuron}(b) as an example. Consider a gait as a periodic solution with period $T$ of the CPG, such that $x_i(t+T)=x_i(t), i=1,2,3,4$.
The spatiotemporal symmetry of gait can be considered as a permutation that preserves the gait rhythm with a phase shift $\theta$.

For a walk gait, the phase relation between neuron 3 and neuron 1 is $x_3(t)=x_1(t+T/4)$, abbreviated as $x_1(t+1/4)$, the phase relation among all neurons are:
\begin{equation}\label{eq:walk_gait}
\begin{array}{c|c}
x_4(t)=x_2(t+\frac{1}{4}) & x_3(t)=x_1(t+\frac{1}{4})\\
\cmidrule(lr){1-2}
x_1(t) & x_2(t)=x_3(t+\frac{1}{4})
\end{array}
\end{equation}
A possible permutation for walk is $\omega(\frac{1}{4})$, where $\omega=(1324)$ mentioned above, $\frac{1}{4}$ is the phase shift.

For trot gait, the phase relation among neurons is:
\begin{equation}\label{eq:trot_gait}
\begin{array}{c|c}
x_4(t)=x_3(t+\frac{1}{2}) & x_3(t)\\
\cmidrule(lr){1-2}
x_1(t) & x_2(t)=x_1(t+\frac{1}{2})
\end{array}
\end{equation}
A possible permutation is $\omega^2(\frac{1}{2})$, which maintains the phase relation between neurons 1 and 2, 3 and 4. Noticed that there is no strict constraint between neurons 1 and 3.

For bound gait, a possible permutation is $\omega(\frac{1}{2})$. The phase relation among neurons is:
\begin{equation}\label{eq:bound_gait}
\begin{array}{c|c}
x_4(t)=x_2(t+\frac{1}{2}) & x_3(t)=x_1(t+\frac{1}{2})\\
\cmidrule(lr){1-2}
x_1(t) & x_2(t)=x_3(t+\frac{1}{2})
\end{array}
\end{equation}

\vspace{0.2cm}
\textit{\textbf{H/K}} \textbf{theorem and calculation}

Consider the symmetry group of the network $\Gamma$, which includes all permutations that preserve the network architecture.
Suppose this network can generate multiple gaits, each gait has a special spatiotemporal symmetry group $H$.
Consider the periodic states of a gait is a periodic solution $x(t)$.
Spatiotemporal symmetries are defined as $\phi \{x(t)\}=\{x(t)\}$, which are permutations that preserve the orbit of the periodic solution, in another word, $\phi \{x(t)\}$ will be the same solution as $x(t)$ but with some phase shift ~\citep{golubitsky_nonlinear_2006, righetti_design_2006}.

There exists a special case where the phase shift is zero after some permutations.
The group composed of these permutations constitutes the spatial symmetry subgroup $K$ of the gait.
Spatial symmetries are defined as $\gamma x(t)=x(t)$, which are permutations that preserve the periodic solution.
The relation among the network symmetry group $\Gamma$, gait spatiotemporal group $H$ and the gait spatial symmetry subgroup $K$ observe the following relations:
$K\subset H \subset \Gamma$.
Moreover, there is another symmetry for gait, that is a cyclic group $H/K$ which contains the permutations that generate phase shift.

Taking the bound gait in the network of Fig.~\ref{fig:4neuron}(b) as an example.
Spatiotemporal symmetry $H$ is $\mathbf{Z}_4=(e,\omega,\omega^2,\omega^3)$.
Spatial symmetry $K$ is $\mathbf{Z}_2(e,\omega^2)$.
To calculate $H/K$, one has to calculate the set of all left cosets of $K$ in $H$:
\begin{equation}\label{eq:H/K calculation bound}
\begin{aligned}
H/K &=\{a\cdot K:a\in H\}\\
    &=\{
    \{e,\omega^2\},
    \{\omega,\omega^3\},
    \{e,\omega^2\},
    \{\omega,\omega^3\}
    \}\\
    &=\{e\cdot\{e,\omega^2\},\omega \cdot \{e,\omega^2\}\}
\end{aligned}
\end{equation}
The result implies $H/K$ is a $\mathbf{Z}_2$ cyclic group. The group elements are $\{e,\omega(\frac{1}{2})\}$.

\section*{Appendix C. Gait transition}
\label{App.Gait transition}

\textbf{Driving signals in \emph{Power Pair} and \emph{Wait \& Power Pair}:}

\begin{figure}[t]
  \centering
  \includegraphics[width= 8 cm]{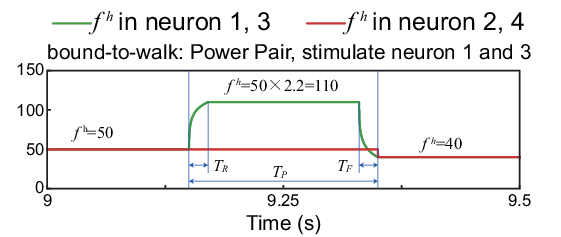}
  \caption{
  The variant of $f^h$  during \emph{Power Pair} and \emph{Wait \& Power Pair}.
  An example of bound-to-walk by \emph{Power Pair} with  $[R_P,T_P,\eta_R,\eta_F]=[2.2,0.2,0.1,0.1]$.
  }
  \label{fig:driving signals}
\end{figure}

In these strategies, the driving signals of the selected neurons are increased for brief periods and then returned to their original values. The parameters $f^h$ in equation~(\ref{eq:driving signals}) are firstly increased for a gain ratio $R_P$ to a time period $T_P$ and then return to the $f^h$ of the target gait. Figure~\ref{fig:driving signals} shows the variation process of $f^h$ during a gait transition.
The duty cycles of the rising edge and the falling edge are $\eta_R=\frac{T_R}{T_P}$ and $\eta_F=\frac{T_F}{T_P}$, respectively.
The shapes of these curves are generated by functions:
\begin{equation}\label{eq:rising edge falling edge}
    \begin{aligned}
    \text{Rising curve}&: y=\log_{10}(x), x\in[1,10]\\
    \text{Falling curve}&: y=\log_{0.1}(x), x\in[0.1,1].
    \end{aligned}
\end{equation}
These primitive curves are then translated and scaled according to parameters.


\vspace{0.2cm}

\textbf{The parameters of the \emph{Power Pair}:}
\begin{itemize}
  \item pronk-to-walk:  $[R_P,T_P,\eta_R,\eta_F]=[2.0, 0.1, 0.1, 0.1]$, stimulate neurons 1 and 3.

  \item pronk-to-bound:$[R_P,T_P,\eta_R,\eta_F]=[2.0, 0.14, 0.1,0.1]$, stimulate neurons 1 and 2.
\end{itemize}

\vspace{0.2cm}

\textbf{The parameters of the \emph{Wait \& Power Pair}:}
\begin{itemize}
  \item Bound-to-walk:\\$[R_P,T_P,\eta_R,\eta_F]=[2.2,0.2,0.1,0.1]$, stimulate neurons 1 and 3.

  \item Bound-to-trot:\\$[R_P,T_P,\eta_R,\eta_F]=[1.8,0.4,0.4,0.4]$, stimulate neurons 1 and 3.

  \item Bound-to-pace:\\$[R_P,T_P,\eta_R,\eta_F]=[2,0.2,0.1,0.1]$, stimulate neurons 1 and 4.

  \item pronk-to-trot:\\$[R_P,T_P,\eta_R,\eta_F]=[2.6,0.07,0.1,0.1]$, stimulate neurons 1 and 3.

  \item pronk-to-pace:\\$[R_P,T_P,\eta_R,\eta_F]=[2.6,0.09,0.1,0.1]$, stimulate neurons 1 and 4.
\end{itemize}

\begin{figure}[t]
  \centering
  \includegraphics[width= 8 cm]{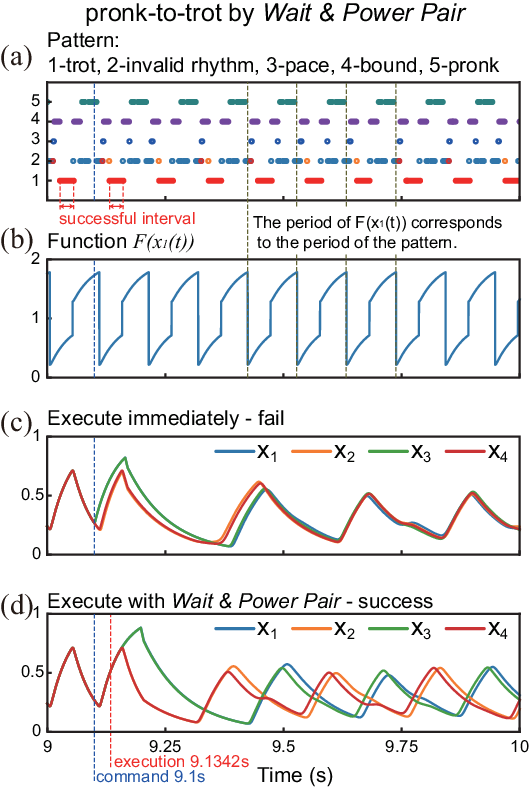}
  \caption{
  Details of applying the \emph{Wait \& Power Pair} strategy in pronk-to-trot transition.
  (a) The pattern of the transition. The trot rhythm after the transition forms the successful interval.
  (b) The function $F(x_1(t))$, the period of the function corresponds to the period of the pattern.
  (c) The transition is executed as the command set, which leads to the failure of the transition.
  (d) The transition is executed with a delay, waiting for the successful interval of the pattern, which leads to the success of the transition.
  }
  \label{fig:wait and powerpair details}
\end{figure}

\textbf{Execution time in \emph{Wait \& Switch} and \emph{Wait \& Power Pair}:}

Some transitions may cause unwanted results including invalid rhythms, unstable transition processes, and wrong target gaits. We further uncovered that the above conditions have fixed patterns referring to the transition execution time (Fig.~\ref{fig:difficulties-in-transitions}). To ensure the success of the transitions, two new strategies \emph{Wait \& Switch} and \emph{Wait \& Power Pair} have been developed based on previous \emph{Wait} and \emph{Power Pair} strategies.

\emph{Wait} operation is designed as follows: after issuing a gait transition command, the controller first checks if the current moment declines within a successful interval of the pattern. If it does, the transition is executed. If not, the controller waits for the system to enter a successful interval of the pattern before executing the transition.

The method for determining whether the current moment declines within a successful interval of the pattern according to the function $F(x_1(t))$.
\begin{equation}\label{eq:Function_neuron1}
F(x_1(t)=\left\{
\begin{aligned}
x_1(t),~~    &\text{if~} \frac{\mathrm{d}x(t)}{\mathrm{d}t}\geq0  \\
2-x_1(t),~~  &\text{if~} \frac{\mathrm{d}x(t)}{\mathrm{d}t}< 0  \\
\end{aligned}
\right.
\end{equation}
This function transforms the signal of neuron 1 into a monotonic function that corresponds to the period of the pattern.

Here, we take the pronk-to-trot transition as an example. As shown in Fig.~\ref{fig:wait and powerpair details}(a) and (b), the period of the function $F(x_1(t))$  corresponds to the period of the pronk-to-trot pattern, and the successful interval of the pattern can be identified through the value of the function.

Fig.~\ref{fig:wait and powerpair details}(c) demonstrates the results of the \emph{Power Pair} strategy without delay. The transition command is given and executed at 9.1s, which leads to the failure of the transition, and the neurons maintain the pronk gait.
Fig.~\ref{fig:wait and powerpair details}(d) demonstrates the results of the \emph{Wait \& Power Pair} strategy with a short delay. The transition command is still given at 9.1s, but the algorithm identifies that the current moment is not in the successful interval through the value of the function $F(x_1(t))$. Thus, it waits until the successful interval starts at 9.1342s and executes the transition, leading to success.

\vspace{0.2cm}

\textbf{The criteria of the \emph{Wait \& Switch}:}
\begin{itemize}
  \item walk-to-trot: $F(x_1(t))\in[1.52,1.7]$

  \item walk-to-pace: $F(x_1(t))\in[0.3,0.56]$

  \item trot-to-pace: $F(x_1(t))\in[1.81,1.84]$

  \item pace-to-trot: $F(x_1(t))\in[1.498,1.615]$

\end{itemize}

\vspace{0.2cm}

\textbf{The criteria of the \emph{Wait \& Power Pair}:}
\begin{itemize}
  \item bound-to-walk: $F(x_1(t))\in[0.56,1.693]$

  \item bound-to-trot: $F(x_1(t))\in[0.58,1.375]$

  \item bound-to-pace: $F(x_1(t))\in[1.82,1.851]$

  \item pronk-to-trot: $F(x_1(t))\in[0.52,0.7]$

  \item pronk-to-pace: $F(x_1(t))\in[1.75,2]$

\end{itemize}

\vspace{0.2cm}

\end{document}